%% file: main.tex
\documentclass[10pt,twocolumn,letterpaper]{article}

\usepackage[pagenumbers]{cvpr}

\input{preamble}

\usepackage{siunitx}
\usepackage{makecell}
\usepackage{graphicx}
\usepackage{multirow}
\usepackage{setspace}
\usepackage{dcolumn}
\usepackage{comment}
\usepackage{colortbl}
\usepackage{pifont}
\usepackage{placeins}
\usepackage[most]{tcolorbox}
\usepackage{fontawesome5}
\definecolor{darkgreen}{rgb}{0.0, 0.5, 0.0}

\usepackage{graphicx}
\usepackage{booktabs}
\usepackage{tabularx}
\usepackage{multirow}
\usepackage{tcolorbox}
\usepackage{pifont}
\usepackage{makecell}
\usepackage{wrapfig}
\tcbuselibrary{listings, breakable}
\usepackage{float}
\usepackage{cuted}
\usepackage{newtxtext,newtxmath}
\definecolor{cvprblue}{rgb}{0.21,0.49,0.74}
\usepackage[pagebackref,breaklinks,colorlinks,allcolors=cvprblue]{hyperref}

\newcommand{\mypar}[1]{\vspace{1mm}\noindent\textbf{#1}}

\newcommand{\cmark}{\textcolor{green!60!black}{\checkmark}}
\newcommand{\xmark}{\textcolor{red!70!black}{\ding{53}}}

\newcommand{\verifierfontsize}{\scriptsize}

\definecolor{boxblue}{RGB}{240, 245, 250}
\definecolor{boxgreen}{RGB}{240, 248, 240}
\definecolor{boxbuff}{RGB}{252, 245, 230}
\definecolor{boxrose}{RGB}{252, 240, 240}
\definecolor{boxpurple}{RGB}{245, 240, 250}
\def\bench{\textsc{StEvo-Bench}\xspace}

\title{Out of Sight, Out of Mind? Evaluating State Evolution in Video World Models}

\begin{document}
\author{
Ziqi Ma\footnotemark[1]
\qquad
Mengzhan Liufu\footnotemark[1]
\qquad
Georgia Gkioxari \\
California Institute of Technology \\
}

\twocolumn[{%
\renewcommand\twocolumn[1][]{#1}%
\maketitle
\begin{center}
    \vspace{-5mm}
    \centering
    \includegraphics[width=0.9\linewidth]{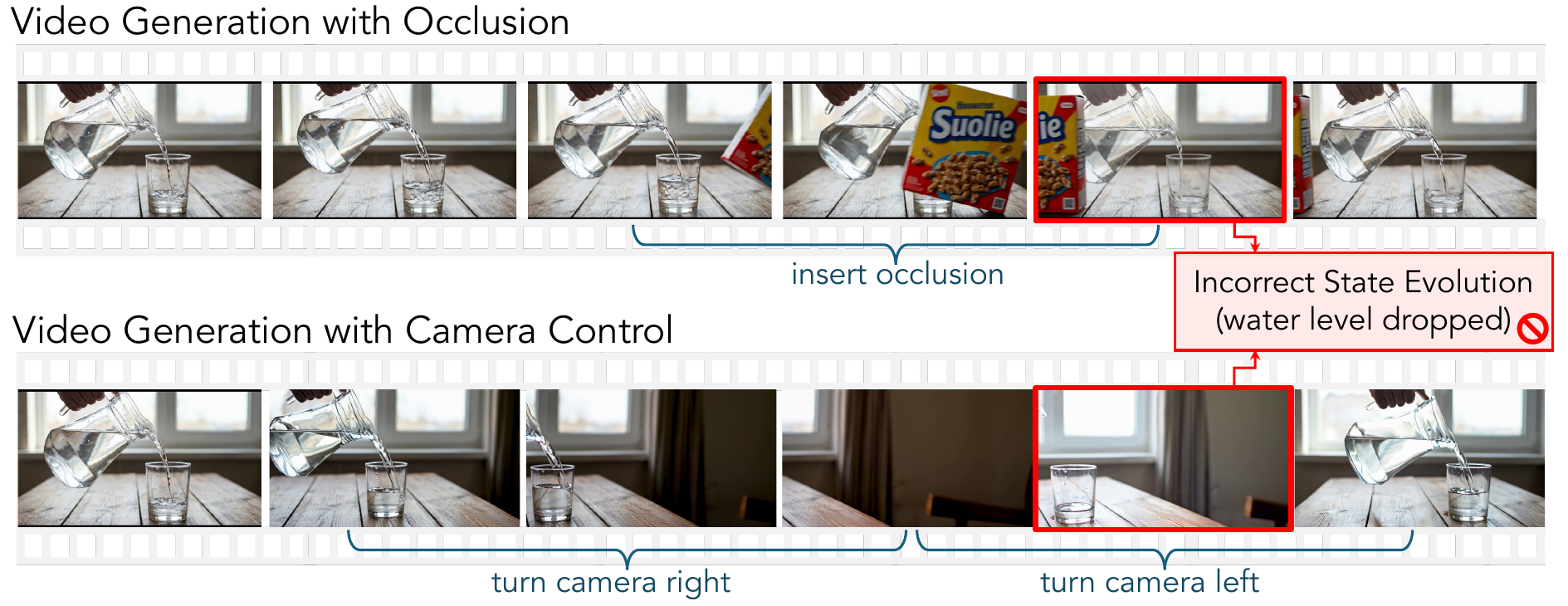}
    \vspace{-2mm}
    \captionof{figure}{Today’s video world models “simulate” the world by generating pixels. We test whether they can separate state evolution from what’s visible by turning the camera away or adding in-scene occlusions. \bench evaluates three key capabilities under lookaway/occlusion: whether evolution continues at all, whether it remains physically plausible, and whether the scene stays coherent.
    }
    \label{fig:teaser}
\end{center}
}]

\footnotetext[1]{Equal contribution.}
\input{secs/00_abstract}
\input{secs/01_introduction}

\input{secs/02_related_works}
\input{secs/03_benchmark}
\input{secs/04_evaluation}

\input{secs/05_analysis}
\input{secs/06_conclusions}

\input{secs/07_acknowledgement}

{
    \small
    \bibliographystyle{ieeenat_fullname}
    \bibliography{main}
}

\input{secs/X_supp}

\end{document}

%% file: preamble.tex
\usepackage{pifont}


%% file: secs/00_abstract.tex
 
\begin{abstract}
Evolutions in the world, such as water pouring or ice melting, happen regardless of being observed. Video world models generate ``worlds'' via 2D frame observations. Can these generated ``worlds'' evolve regardless of observation? To probe this question, we design a benchmark to evaluate whether video world models can decouple state evolution from observation. Our benchmark, \bench, applies observation control to evolving processes via instructions of occluder insertion, turning off the light, or specifying camera ``lookaway'' trajectories. By evaluating video models with and without camera control for a diverse set of naturally-occurring evolutions, we expose their limitations in decoupling state evolution from observation. \bench proposes an evaluation protocol to automatically detect and disentangle failure modes of video world models across key aspects of natural state evolution. Analysis of \bench results provide new insight into potential data and architecture bias of present-day video world models. Project website: \url{https://glab-caltech.github.io/STEVOBench/}. Blog: \url{https://ziqi-ma.github.io/blog/2026/outofsight/}

\end{abstract}


%% file: secs/01_introduction.tex
\section{Introduction}

\begin{figure*}[t!]
\centering
\includegraphics[width=\linewidth]{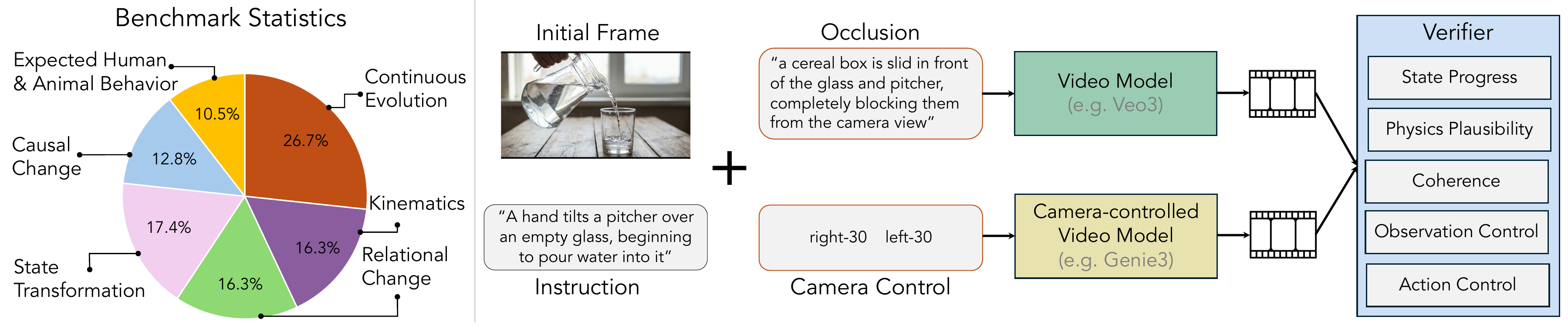}
\vspace{-6mm}
\caption{\bench probes whether video-based world models can decouple state evolution from observation. \bench, consisting of 225 unique tasks spanning 6 categories, uses (image, text prompt, camera control) tuples to prompt video-based world models to generate evolutions under interrupted observation. Observation interruption is done either using text prompt (such as adding occluders, turning off the light), or using camera control (turning the camera view away). The generated videos go through \bench's automatic verifiers for evaluation.}
\label{fig:benchmark}
\end{figure*}

The world evolves regardless of whether we observe it or not.
Consider~\cref{fig:teaser}. When water is being poured into a cup, the amount of water grows, regardless of whether the cup is observed.
Fueled by the rapid progress in video generation, today's world models can generate visual worlds via synthesizing image frames.
These image frames, by depicting visual appearance, represent a world with objects and properties, forming the ``world state''. As objects move or properties change, such as humans moving or fire burning, the ``world state'' evolves. In the real world, we know that these state evolutions are not dependent on observation -- even if they are occluded, or out of view, the state evolves as expected.
If video models are truly ``world models'', they need to possess this property as well: states should correctly evolve, even if not observed.

This is not merely an intellectual investigation. If we want our world models to generate larger worlds and enable longer-horizon interactions, the visible frame observation for any given moment becomes an increasingly small fraction of the generated world. In other words, the majority of the generated world is unobserved. Given this, it becomes critical that the world models can continue evolving the world state, decoupled from observation.

Decoupling evolution from observation has three implicit aspects: 1) The physical plausibility of state evolution: For example, does the water flow correctly based on gravity, and do glass containers not penetrate each other? 2) Coherence of non-evolving aspects: For example, do the cup and jug have the same material and size across the video, without abrupt change or disappearance? 3) The progress of state evolution while not being observed: For example, when the camera turns away and back while the water is still pouring, does water level grow?

Benchmarks define the goal posts for model development, yet existing ones do not support this critical aspect of video world models. Prior benchmarks of world models touch upon 1) and 2). For example, intuitive physics benchmarks like \cite{bansal2024videophy, bansal2025videophy2, mak2026physicsmindsimrealmechanics} focus on 1) for physical processes. Consistency benchmarks like \cite{wu2026considgenviewconsistentidentitypreservingimagetovideo}, or consistency subcategories of \cite{huang2023vbench, zheng2025vbench2, duan2025worldscore}, focus on 2) under full observation. Memory benchmarks like \cite{ye2026mindbenchmarkingmemoryconsistency} involve less-than-full observation, but only focus on coherence for static settings. Current benchmarks, even when combined, only cover 1) and 2), and thus cannot support the growth of world models towards bigger worlds, longer horizon, which naturally introduces more processes that cannot be observed in their full duration.

We present a benchmark which evaluates all three aspects to verify whether video world models can decouple state evolution from observation. Our benchmark, \bench, inserts occlusions or directs the camera to look away during an evolving process. For example, we let the model generate a burning match, and then either insert an occlusion in front of the camera or control the camera to turn away while the match is burning. When the camera turns back or the occlusion is removed, we evaluate whether the match has burned more than previously seen. We focus on video world models, which include image(text)-to-video models and camera-controlled video models.
\bench contains 225 tasks, encompassing six evolution categories: continuous process, kinematics, relational change, causal change, state transformation, and expected human/animal behavior. To enable automatic evaluation, \bench also introduces specialist verifiers built upon VLMs, which automatically evaluate and disentangle critical failure modes.

\bench allows us to probe how world models handle evolving processes when observation stops, exposing interesting findings about these models. For video models \cite{veo2025, sora2025, wan2025, kong2024hunyuanvideo, yang2025cogvideox}, we find that general-purpose video generation models exhibit evolution stopping or incoherence when observation control (such as occlusion or turning off the light) is applied. For camera-controlled models \cite{genie2025, lingbot-world, sun2025worldplay}, we notice a strong bias toward static scenes (no evolution) and increased difficulty in camera control when the models are able to evolve state. We further find that memory-based architectures exacerbate the static-scene bias, and does not help decouple evolution from observation. These findings motivate further investigations about data and architecture bias of current video world models.

We make the following contributions:
\begin{itemize}
\item We propose \bench, a benchmark to measure whether world models can decouple state evolution from observation.
\item We build automatic specialist verifiers to detect and disentangle various axes of failure in state evolutions generated by video world models.
\item Quantitative analysis on \bench reveals critical findings about current state-of-the-art video world models.
\end{itemize}

%% file: secs/02_related_works.tex
\section{Related Works}
\input{tables/benchmark_scope}

\mypar{Video World Models.}
With the rapid progress in video generation, people started to term video models ``world models'' due to their capability to generate realistic-looking worlds via video frames. Video ``world models" can be categorized into two types: The first type is general-purpose video generation models, such as \cite{veo2025, sora2025, wan2025, kong2024hunyuanvideo, zheng2026opensora20trainingcommerciallevel, yang2025cogvideox}, which can generate worlds based on image or text prompts. The other type is action-conditioned video models \cite{genie2025, sun2025worldplay, lingbot-world, bar2025navigation}, which generate videos semi-autoregressively, conditioned on actions. Many works focus on camera control (i.e. navigation) as action \cite{genie2025, lingbot-world, sun2025worldplay, bar2025navigation}. Action space could also be robotic end effector space \cite{ali2025world}, or time plus camera control \cite{wang2025bullettime}. While the definition of ``world models'' can be broad, in this work, we focus on general-domain world models of the two types above.

\mypar{World Models Evaluation.}
Early evaluations of video models focus on video quality and general condition-consistency \cite{huang2023vbench}. However, using video models to generate realistic ``worlds'', rather than just visually-pleasing videos, poses additional requirements, such as physics correctness, common sense, controllability, and dynamics. General world model benchmarks such as WorldScore \cite{duan2025worldscore}, WorldModelBench\cite{Li2025WorldModelBench} test these aspects in simple settings. Among specialized benchmarks, VideoPhy \cite{bansal2024videophy}, VideoPhy2 \cite{bansal2025videophy2} and PAIBench \cite{zhou2025pai} focus on physics realism, World Consistency Score \cite{rakheja2025worldconsistency}, Stable World \cite{kwon2026toward} and MIND \cite{ye2026mindbenchmarkingmemoryconsistency} focus on consistency and memory. In this work, we evaluate whether world models can successfully evolve processes without full observation, presenting a more rigorous test of physical plausibility and coherence compounded with the challenges of interrupted observation.

\mypar{State in World Models.}
Some prior works propose a ``stateful'' world model that carries an implicit world state representation. Such a formulation, by definition, decouples the state's evolution from observation. Prior works in this direction can be categorized by their different state representations: memory as state, 2.5D/3D as state, and latent state. WorldMem \cite{xiao2025worldmem}, VMem \cite{li2025vmem}, and Long-term Spatial Memory \cite{wu2025longtermmemory} represent the world state as memory. WonderPlay \cite{li2025wonderplay}, VerseCrafter \cite{zheng2026versecrafter}, and Point World\cite{huang2026pointworld} use depth point cloud or 3D representations as the ``world state''. These states, either memory-based or 3D-based, strongly bias towards static scenes. Finally, works like Dino-WM \cite{zhou2025dinowm}, V-JEPA2 \cite{assran2025vjepa}, Long-context SSM \cite{po2025long}, and FloWM \cite{lillemark2026flow} use a latent world state. FloWM discusses latent state evolution in simple, synthetic environments like block world. While prior works on stateful world models are generally constrained in simple, synthetic environments, we design a method to probe the ``statefulness'' in general-purpose video-based world models.

%% file: tables/benchmark_scope.tex
\begin{table*}[h!]
\centering
\resizebox{1.8\columnwidth}{!}{
\begin{tabular}{lccc @{\hskip 15pt}cc}
\toprule
& \multicolumn{3}{c}{\textit{Evolution criteria}} & \multicolumn{2}{c}{\textit{Control criteria}} \\
\cmidrule(lr){2-4} \cmidrule(lr){5-6}
\multicolumn{1}{c}{Category} & Progress & Physics & Coherence & Observation & Action \\
\midrule
Physics \& Common sense \cite{Li2025WorldModelBench, bansal2024videophy, bansal2025videophy2,li2025smallworlds}
  & \cmark & \cmark & \xmark & \xmark & \xmark \\
Memory \& consistency \cite{ye2026mindbenchmarkingmemoryconsistency, rakheja2025worldconsistency, wu2026considgenviewconsistentidentitypreservingimagetovideo}
  & \xmark & \xmark & \cmark & \cmark & \xmark \\
Instruction following \cite{zheng2025vbench2, duan2025worldscore}
  & \xmark & \xmark & \xmark & \xmark & \cmark \\
\bench
  & \cmark & \cmark & \cmark & \cmark & \cmark \\
\bottomrule
\end{tabular}
}
\vspace{1mm}
\caption{Evaluating state evolution decoupled from observation requires multiple criteria. While prior benchmarks for video (world) models only touch on subsets of them,, \bench comprehensively evaluates all aspects to expose and disentangle failure modes of world models when generating evolutions that are not fully observed.}
\label{table:benchmark_scope}
\vspace{-3mm}
\end{table*}

%% file: secs/03_benchmark.tex
\section{Benchmarking State Evolution in Video World Models}

\begin{figure*}[t!]
    \centering
    \vspace{-3mm}
    \includegraphics[width=\linewidth]{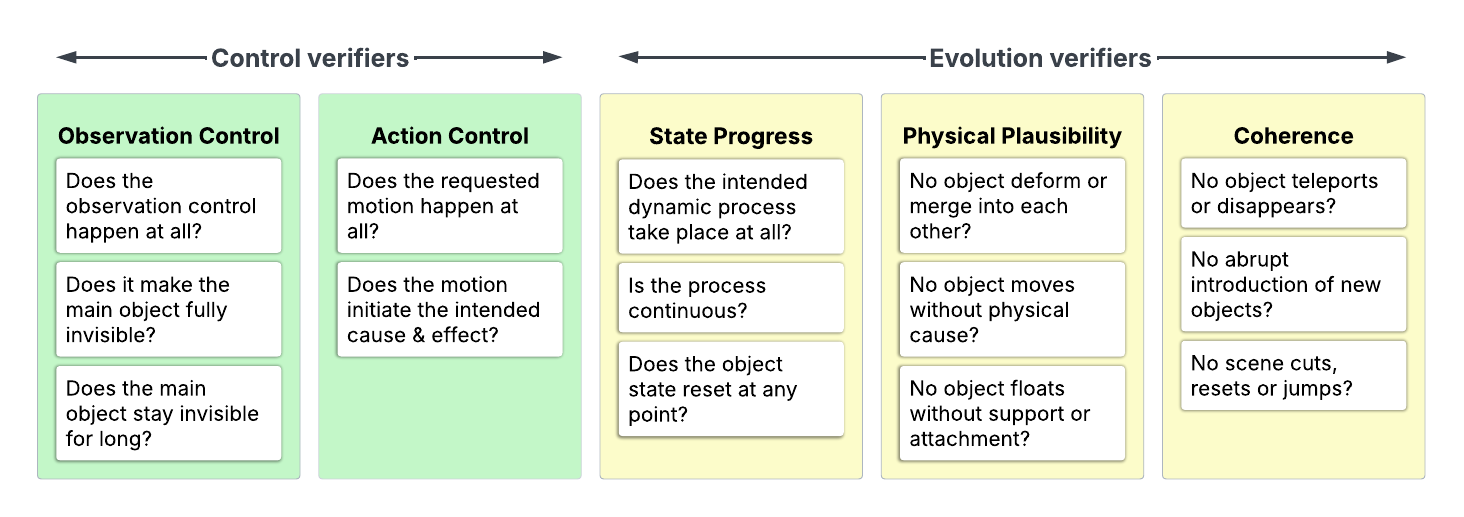}
    \vspace{-6mm}
    \caption{Automatic verifier pipeline for \bench. Five video understanding verifiers independently assess each generated video on one criteria. Observation Control and Action Control jointly determine Control Success; State Progress, Physics Plausibility, and Coherence jointly determine Evolution Success.}
    \label{fig:verifier}
    \vspace{-3mm}
\end{figure*}

\bench evaluates whether video world models can continue scene evolution when not the whole process is in view. It consists of 225 unique tasks, spanning 6 different categories of naturally-occurring evolutions, including continuous process, kinematics, relational change, causal change, state transformation, and expected human/animal behavior. The tasks reflect physical events that real-world agents routinely observe: a hand turns on a stove burner and a wet pan begins to heat, a pitcher pours water into a glass, a wall switch is toggled and a lamp turns on, a tire pump inflates a deflated tire, a railroad crossing gate descends as a train approaches, or a hand turns a manual valve and a sprinkler activates. The diversity of tasks and their grounding in everyday physical interactions make \bench directly relevant to the challenges faced by embodied agents deployed in the real world.

\subsection{Task Construction}

Tasks in \bench require world models to generate a continuous process evolution under interrupted observation (\cref{fig:benchmark}). Each task is specified by an initial image and a text prompt. The initial image shows a main object in its initial state. The text prompt instructs the world model to apply two \emph{controls}: an \emph{action control} that initiates an evolution process of the main object (\eg, hand tips over the first domino), and an \emph{observation control} that interrupts observation of the process mid-evolution. We evaluate the state of the main object before and after observation of it is interrupted.

We implement observation control in a model-specific way. For video generation models, the text prompt instructs the model to apply an in-scene observation interruption for some prolonged period and then remove it to reveal the main object again. That is either a physical occluder placed in front of the camera (\eg a piece of cardboard or curtains) or removing all illumination (\eg turning lights off). For camera-controlled video models, we specify a camera trajectory that moves the main object fully out of frame and then returns to bring it back in view (\eg, ``right-30 steps, left-30 steps''). In both cases, the final revealed view indicates whether the state of the main object has evolved correctly during the interval it was out of sight. These strategies are tailored to each model's control interface to ensure reliable observation interruption.

\subsection{Evaluation Criteria and Automatic Verifiers}
\label{sec:metrics}

\bench evaluates generated videos using a two-stage pipeline. First, we check \emph{control success}: whether the model applied both controls — the observation control successfully hid the scene, and the action control successfully initiated the process. Tasks that fail these control criteria are excluded from further evaluation, since without successful controls the process is either uninitiated or not partially observed. On the passing subset, we evaluate \emph{task success}, which requires three evolution criteria to hold: \emph{state progress} (did the process qualitatively occur during occlusion?), \emph{physical plausibility} (is the evolution physically correct?), and \emph{coherence} (does the video remain temporally consistent before and after occlusion?). A video achieves task success only if it first achieves control success and then passes all three evolution criteria.

Previous benchmarks focus on a subset of these aspects, as shown in \cref{table:benchmark_scope}. \bench evaluates all these aspects to comprehensively quantify state evolution success and disentangle various failure modes.

To evaluate these criteria automatically, we build a specialist verifier that produces a single binary judgment (\cref{fig:verifier}). All verifiers use \texttt{Gemini 3.1 Pro} as the VLM judge. Decomposing verification into independent specialists serves two purposes: (1) It enables fine-grained diagnosis of why a model fails, since current video models often exhibit multiple entangled failure modes in a single generation; (2) Each verifier poses a single, narrowly scoped yes/no question, which produces more reliable VLM judgments than a monolithic prompt that asks the VLM to simultaneously weigh and disentangle multiple aspects of success. This observation is consistent with findings in prior work on checklist-style prompting \cite{viswanathan2025checklist}. 

\paragraph{Control verifiers.}
The two control verifiers check whether the model correctly applied the requested controls:

\mypar{Observation control verifier:} verifies whether the scene was successfully hidden during the evolution, either by an in-scene occluder (cardboard, curtains, or lights off) or by a camera lookaway. The verifier checks that the main object was completely blocked from view for a sustained duration while the process was underway. Partial or transient occlusion, and occlusion that occurs only after the evolution has already completed, are both considered unsuccessful. 

\mypar{Action control verifier:} verifies that the requested action event both occurred and took effect. This requires the motion itself to be visible, and it must also correctly initiate the expected physical response. In the case of the requested action being ``finger tips over the first domino'', the tipping motion must occur, and it must cause the first domino to tilt over. A completely absent tipping motion, or tipping in mid-ar, are both consider failures by this verifier. 

\paragraph{Evolution verifiers.}
If a generated video passes the control verifiers, then three evolution verifiers check whether the world state evolved correctly in the video:

\mypar{State progress verifier:} verifies whether the intended state change progresses at all during occlusion or lookaway, regardless of physical accuracy or completeness. The key question here is whether the process evolves continuously, as opposed to (a) remaining completely static, (b) discontinued during the unobserved period, or (c) having undone itself during the unobserved period. This verifier first uses the VLM's reasoning to predict, in language, what directional change the evolution should produce, then uses a unanimous-vote ensemble ($n=3$) of video understanding models to decide whether that change progressed correctly during the unobserved period.

\mypar{Physical plausibility verifier:} verifies that the evolution is physically correct. The verifier prompt covers a two-type taxonomy of violations: Type 1 covers instantaneous single-frame violations (\eg, rigid objects deformed without force, objects floating against gravity); Type 2 covers dynamic violations where a cause is shown but the wrong effect follows (e.g., smoke sinks downward instead of rising). We prompt the verifier to flag only active violations, not the simple lack of change, to avoid conflating a missing evolution with an incorrect one. The same majority-vote ensemble approach was used to suppress hallucinations

\mypar{Coherence verifier:} verifies that the video is temporally consistent before and after occlusion, both in the main object and the broader scene. Each instance in a majority-vote ensemble ($n=3$) applies a structured checklist: Does the main object vanish with no physical cause? Does it teleport to an unexplained position? Does any new object appear abruptly? Does the scene cut, reset, or resume in a noticeably different configuration? The video passes only if none of the above occurs.

%% file: secs/04_evaluation.tex
\section{Do Video World Models Evolve State Successfully?}
\bench allows us to probe whether video models can successfully evolve states when part of the dynamic process is occluded or out-of-view.

\subsection{Model Candidates}
We evaluate both video models which generate videos conditioned on image and text, as well as camera-controlled video models which additionally takes in a specified camera trajectory.

\mypar{Video Models.}
We evaluate open- and closed-source general-purpose video generative models, including Veo 3 \cite{veo2025}, Sora 2 Pro \cite{sora2025}, WAN 2.2 \cite{wan2025}, CogVideoX 1.5 \cite{yang2025cogvideox} and HunyuanVideo 1.5 \cite{kong2024hunyuanvideo}.

\mypar{Camera-controlled Video Models.}
We evaluate state-of-the-art camera- controlled video models, including Genie 3 \cite{genie2025}, HunyuanWorld 1.0  \cite{sun2025worldplay}, LingBot-World \cite{lingbot-world}, GEN3C \cite{ren2025gen3c}, VMem \cite{li2025vmem}, and AETHER \cite{zhu2025aether}. Among these models, Hunyuan-WorldPlay, Lingbot are open-source general camera-conditioned video models. GEN3C incorporates 3D priors, VMem incorporates a memory module, and AETHER pretrains on 4D reconstruction objective. Since VMem always creates severe visual artifacts during camera pan, and AETHER ``forgets'' the object before camera turn due to its short, 41-frame context, we only evaluate them qualitatively.

\subsection{Findings}
\input{tables/comp_models}

\mypar{Can video world models decouple state evolution from observation?}
\cref{table:metrics_all} quantitatively evaluates whether video models can successfully evolve states on \bench tasks, as well as their subgoal performance.

\begin{tcolorbox}[colback=blue!5!white, colframe=blue!75!black]
  Finding 1: Both video models and camera-controlled video models struggle to decouple state evolution from observation. 
\end{tcolorbox}

As seen in \cref{table:metrics_all}, both video models and camera-controlled video models show less than $10\%$ success rate in evolving state under interrupted observation. While closed-source video models such as Veo 3 and Sora 2 Pro can occasionally evolve the process (state progress rate being $17\%$ and $13\%$ respectively), they often fail to maintain coherence and physical plausibility. Camera-controlled models almost never evolve states, with less than $5\%$ state progress across closed- and open-source models. We present findings about these models' specific failure modes in the analysis below.

\begin{figure*}[t!]
\centering
\includegraphics[width=\linewidth]{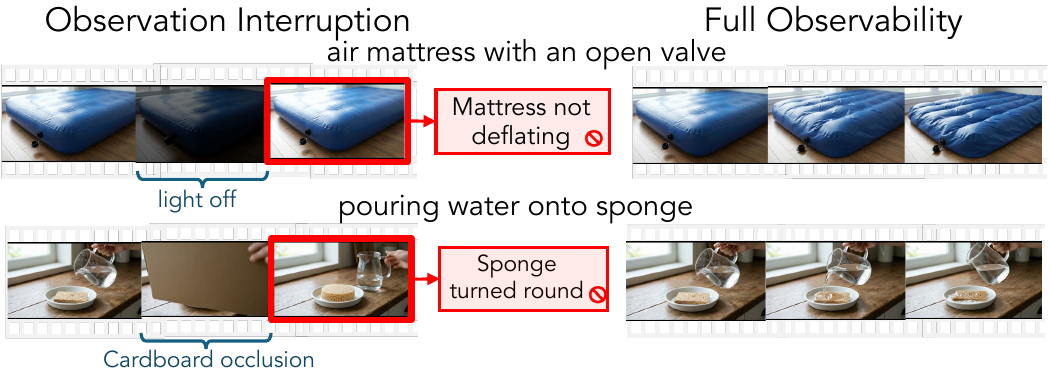}
\caption{Representative failure modes of video models when observation of the evolution process is interrupted. The model is capable of simulating the process correctly when the scene is fully visible. However, when observation of the process is temporarily interrupted, the model either stops evolving state, as seen in the mattress deflation example, or fails to preserve object coherence, as seen in the sponge example. These videos are generated by Veo 3.}
\label{fig:video_failure}
\end{figure*}

\mypar{How do video models fail under observation control?}
\bench exposes two key failure modes of video models when observation control is applied to these models: evolution stopping and incoherence. \cref{fig:video_failure} highlights both failure modes. 

\begin{tcolorbox}[colback=blue!5!white, colframe=blue!75!black]
  Finding 2: 
  When observation control (\eg, occlusions or “turn light off”) is introduced into an evolving process, video models often stall the evolution or exhibit incoherence.
\end{tcolorbox}

\paragraph{Evolution stopping.} 
A common failure mode is stalled state evolution. For instance, in \cref{fig:video_failure}, an air mattress with an open valve should continue deflating; yet when we instruct the model to “turn off the light”, the mattress stops deflating. Our state progress verifier detects this issue. As shown in \cref{table:metrics_all}, all video models have low state progress rates, with open-source models particularly prone to this failure mode.

\paragraph{Incoherence.} A second common failure mode is loss of object coherence once an occluder disappears -- for example, the rectangular sponge becoming round in \cref{fig:video_failure}. The coherence verifier captures these errors, with state-of-the-art closed-source models achieving only about a 60\% success rate.

\input{tables/partial_full_compare}

\begin{figure*}[ht!]
\centering
\includegraphics[width=\linewidth]{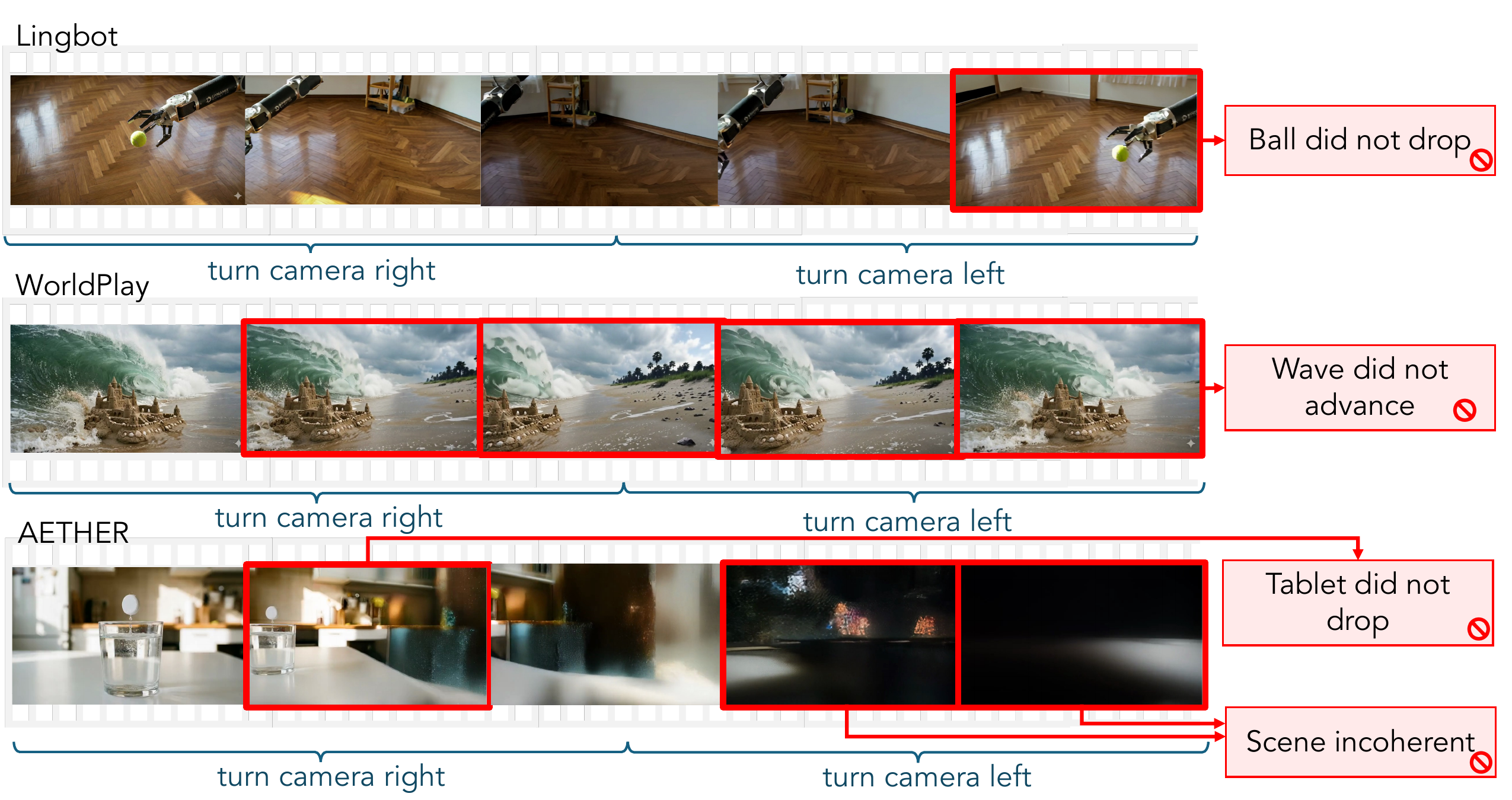}
\caption{Open-source camera-controlled video models assume the scene to be static as the camera turns, and fail to correctly evolve state, such as the ball dropping, wave advancing, or tablet dropping.}
\label{fig:camera_failure}
\end{figure*}

To further understand whether such failures are indeed caused by the observation control, we perform additional experiments where the observation control is not inserted, \ie, the process evolves while fully observed. As seen in \cref{fig:video_failure}, both these processes succeed when there is no occluder or ``light off'' intervention. \cref{table:video_compare_full} provides further quantitative evidence that when observation control is applied, both the state progress rate (which measures whether the state progressed during the occlusion) and task success rate (which measures whether the full evolution is correct and coherent) are significantly lower.

The success of full-observability generation suggests that models possesses internal knowledge of how the process should evolve. However, observation control renders these models incapable of generating these processes correctly. 
While one might argue that this problem is solvable by incorporating large-scale occlusion-interrupted dynamics videos into training, we believe such failures expose fundamental issues on how video models process context. Since the occluding frames do not provide any useful information on the state evolution, naive all-to-all bidirectional attention might not be efficient in such cases. We hope these failure cases can inspire and motivate new architecture designs for video world models.

\mypar{How do camera-controlled video models fail under observation control?}
\begin{tcolorbox}[colback=blue!5!white, colframe=blue!75!black]
 Finding 3: Camera-controlled video models tend to ``freeze'' the scene and fail to continue the dynamic process.
\end{tcolorbox}

\begin{figure*}[h!]
\centering
\includegraphics[width=\linewidth]{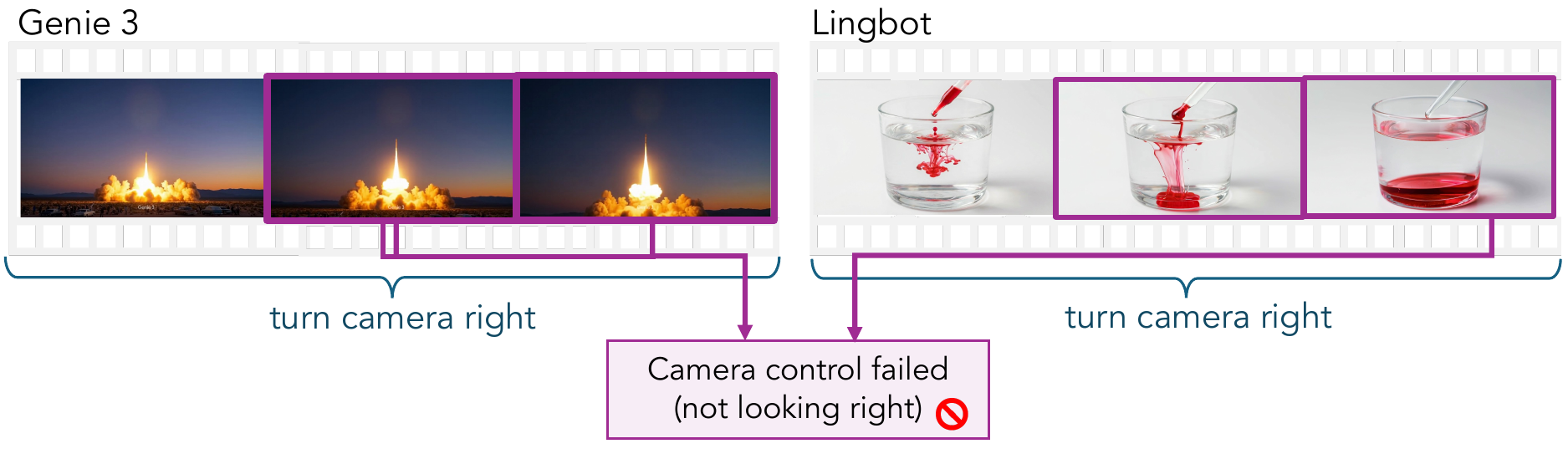}
\caption{When camera-controlled models can successfully evolve states, the model tends to ignore camera control while the evolution is happening. Such behavior is seen both in Genie 3 and Lingbot.}
\label{fig:genie_tradeoff}
\end{figure*}

Camera-controlled models, when instructed to ``look away'' from a process, largely fail to evolve the process, and tend to keep the scene static as the camera moves. We observe similar behavior across models, especially open-source models like Hunyuan-WorldPlay, Lingbot, and GEN3C. This is also seen in the close-to-zero state progress rate for all camera-controlled models in \cref{table:metrics_all}. \cref{fig:camera_failure} shows representative failures of such models, sourced from Hunyuan-WorldPlay and Lingbot. GEN3C, which leverages 3D cache to enhance camera control, biases strongly toward static scene generation. AETHER, a camera-controlled video which mixes in 4D reconstruction objective during training, is also unable to evolve state, and fails to remember the object due to the its short context window.

\begin{figure*}[h!]
\centering
\includegraphics[width=0.8\linewidth]{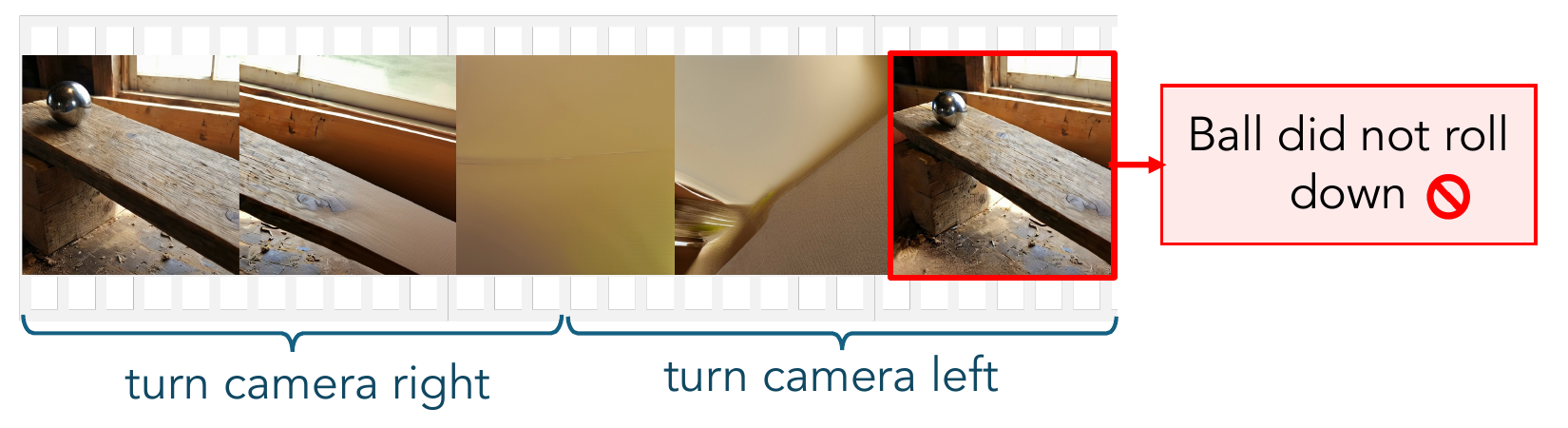}
\caption{Qualitative example of VMem. Due to the memory design, VMem can perfectly recall the initial frame, but fails to evolve state, and suffers from scene incoherence.}
\label{fig:vmem}
\vspace{-6mm}
\end{figure*}

\begin{tcolorbox}[colback=blue!5!white, colframe=blue!75!black]
 Finding 4: When camera-controlled video models successfully evolve the state, camera control becomes difficult.
\end{tcolorbox}

The previous finding states that camera-controlled models frequently generate static worlds. In the less frequent event where they do generate dynamics, we observe that camera control becomes difficult. When state evolution is being generated, the model largely ignores the camera control and keeps the camera static. \cref{fig:genie_tradeoff} shows two representative examples. In both cases, the user continuously directs the camera to turn right. In the first example, while the rocket launches, the camera does not turn right but slightly turns up to follow the rocket. In the second example, the camera stays completely static despite the camera trajectory input. While previous findings focus on whether observation control makes evolution difficult, these failure modes show the converse is also true: evolution makes camera control difficult. 

These behaviors show another aspect of the evolution-observation coupling: the inability for camera pan and state evolution to happen concurrently suggests a strong coupling between the model's capability to evolve state and generating full pixel descriptions of the process from a largely unchanged viewpoint.

We hypothesize that both the camera-controlled models' bias towards static scene and the tradeoff between dynamics and camera control they exhibit might be attributed to the training data. These models usually train on a combination of general video data, rendered 3D data, and gaming data:
\begin{itemize}
\item In order to obtain diverse camera trajectories, camera-controlled models train on rendered videos of static scenes. For example, WorldPlay trains on renderings of reconstructed 3D Gaussian Splats, and LingBot trains on scene renderings from Unreal Engine. These videos move the camera but do not contain dynamics, which might reinforce the static bias, and furthermore associate complex camera movement with static scenes.

\item The general video datasets that models train on, \eg, Sekai \cite{li2025sekaivideodatasetworld} and DL3DV \cite{ling2024dl3dv}, mainly contain indoor and outdoor scenes with many static objects.

\item The gaming data, which contain (action, video) pairs, usually don't contain rich, natural physical evolutions that occur in the real world.
\end{itemize}

Enabling simultanous state evolution and camera control in world models might require different synthetic data strategies to mitigate data bias or post-training with preference optimization.

\mypar{Do memory modules help decouple state evolution from observation?}
Previous findings have shown that video world models have strong coupling between evolution and observation. We further study whether memory modules, which enable the model to keep track of a ``memory state'' separate from pixel-level generations, can help decouple state evolution from observation.

While many memory-based architectures \cite{zhao2025spatia, xiao2025worldmem} are constrained to specialized domains  such as Minecraft or RealEstate10k~\cite{realestate10k}, we evaluate VMem \cite{li2025vmem}, which is the most general-domain memory-based model, trained on indoor and outdoor scenes.

\begin{tcolorbox}[colback=blue!5!white, colframe=blue!75!black]
 Finding 5: Memory-based video world models reinforce the bias toward static scene.
\end{tcolorbox}

As shown in \cref{fig:vmem}, VMem shows a strong bias toward remembering objects exactly ``as is''. VMem, although producing artifacts during the camera lookaway process, is able to recall the initial frame quite accurately due to the memory design. We argue that while memory enables the model to store a ``state'' that is decoupled from observation, the architecture design encourages memorization of object appearance, rather than enabling better modeling of state evolution. \bench highlights the need for new architecture designs to enable not only memorization, but also evolution of the world that is not always fully in-view.

%% file: tables/comp_models.tex
\begin{table*}[t!]
\centering
\begin{tabular}{p{0.07\textwidth}l|cccc}
\toprule
&\centering Model & Success & Progress & Physics & Coherence \\
\midrule

 \multirow{5}{*}{\rotatebox{90}{\shortstack[l]{Video\\models}}}
 & Veo 3          & \textbf{8.7}  & \textbf{17.4} & 82.6 & 66.5 \\
 & Sora 2 Pro     & 8.1  & 13.1 & \textbf{85.5} & \textbf{69.7}  \\
 & WAN 2.2        & 0.9  & 7.7 & 52 & 58.4  \\
 & HY-Video 1.5 & 0.9 & 4.1 & 42.1 & 59.1 \\
 & CogVideoX 1.5  & 0.5  & 1.4 & 68.5 & 67.1  \\
\midrule
\addlinespace[1pt]
 \multirow{5}{*}{\rotatebox{90}{\shortstack[l]{Camera-\\controlled}}}
& Genie 3         & 0.0 & 2.9 & 15.2  & 27.3 \\
& HY-WorldPlay & 0.0 & 0.0 & \textbf{72.2} & \textbf{88.2}  \\
& Lingbot        & 0.0 & \textbf{3.4} & 40.7 & 76.3  \\
& GEN3C          & 0.0 & 0.0 & 30.6 & 82.4 \\
\addlinespace[3pt]
\bottomrule
\end{tabular}
\caption{Evaluation metrics for video models (top) and camera-controlled models (bottom). Both video models and camera-controlled video models struggle to evolve state correctly when observation control is applied. Metrics are in percentage (\%).}
\label{table:metrics_all}
\end{table*}

%% file: tables/partial_full_compare.tex
\begin{table}[h]
\centering
\begin{tabular}{l|cc}
\toprule
    & {\small Progress } & {\small Success } \\
\midrule
Observ. Full    & 84.6   & 46.2   \\
Observ. Control & 17.4   & 12.4    \\ 
\bottomrule
\end{tabular}
\caption{The effect of observation control: state progress rate and task success rate drop significantly when observation control is applied. Metrics are averaged over Veo 3 and Sora 2 Pro.}
\label{table:video_compare_full}
\end{table}

%% file: secs/05_analysis.tex
\input{tables/verifier_agreement}
\section{Control and Verifier Analysis}

\input{tables/control_metrics}

\subsection{Control success rate}


One predicate of our evaluation is that both observation and action control are applied correctly, \ie, the process is correctly initiated, and occlusion or lookaway happen. We verify that current models can support such control by evaluating success rate under each axis. \cref{table:controllability} shows that video models and camera-controlled video models can correctly apply both observation and action control, validating the evaluation protocol of \bench.

\subsection{Evaluation of verifiers against human annotators}

The automatic verifiers are only useful if they reliably capture the intended evaluation criteria. Since visually judging diverse videos is a subjective task, we validate the verifiers by comparing them to human annotations. If a verifier agrees with human annotators at least as well as annotators agree with each other, it is performing at the ceiling of what any automatic system could reasonably achieve. We recruited three crowdsourced annotators on Upwork and collect binary labels (yes/no) for all five evaluation criteria. We do this on a randomly selected subset ($n=180$) of the generated videos spanning all models.

We measure agreement using three complementary metrics. \texttt{Accuracy} is the fraction of tasks where two raters assign the same binary label, averaged over all rater pairs. \texttt{ROC-AUC} addresses potential label class imbalance. Given two raters, one is treated as ground truth and the other as predictor. Since predictions are binary (not soft scores), the \texttt{ROC} curve has only one interior point, so AUC reduces to $(\texttt{TPR} + \texttt{TNR}) / 2$, where \texttt{TPR} is the fraction of ground-truth positives the predictor labels positive, and $TNR$ is the fraction of ground-truth negatives the predictor labels negative. For verifier-vs-human comparisons, the human is treated as ground truth. For human-vs-human comparisons, both directions are computed for each pair and averaged symmetrically, since neither annotator is privileged. \texttt{Model ranking agreement (MRA)} sidesteps absolute calibration entirely: for every pair of models $(A, B)$ and every shared task, it asks whether two raters agree on which model performed better. Agreement is scored as 1.0 for a full match, 0.5 when one rater calls a tie and the other does not, and 0.0 for a reversal; scores are averaged over all model pairs and tasks.

All three metrics produce values between 0 and 1, with 1 being perfect agreement. The agreement results are shown in~\cref{tab:verifier_agreement}. All three metrics converge on consistent conclusions. Human agreement is lowest on physical plausibility and observation control. This reflects genuine ambiguity in the definitions of these conflated, inter-related failure modes. Deciding whether a physical effect constitutes a jarring violation (rather than a slow or imperfect evolution) requires nuanced domain knowledge, and different annotators apply different thresholds. Similarly, judging whether an occlusion was ``complete enough'' is subjective.

Nevertheless, the verifier matches or exceeds inter-human agreement on nearly all criteria. The verifier-vs-human \texttt{MRA} meets or exceeds inter-human \texttt{MRA} on every single criteria. With \texttt{Accuracy}, the verifier outperforms inter-human agreement when evaluating state progress, observation control and action control. The only criteria where the verifier lags slightly is physical plausibility, which is also the hardest for humans. Together, these results confirm that our automatic verifiers are reliable proxies for well-educated human judgment: they agree with individual human annotators as well as annotators agree with each other, and they recover the same model ordering that a human panel would produce.

%% file: tables/verifier_agreement.tex
\begin{table*}[t]
  \centering
  \resizebox{0.8\textwidth}{!}{%
  \begin{tabular}{l @{\hskip 8pt} c cccc @{\hskip 15pt} ccc}
  \toprule
  & & \multicolumn{4}{c}{\textit{Evolution criteria}} & \multicolumn{3}{c}{\textit{Control criteria}} \\
  \cmidrule(lr){3-6} \cmidrule(lr){7-9}
  & \small Metric & \small Progress & \small Physics & \small Coherence & \small Success & \small Observation & \small Action & \small Success \\
  \midrule
  \multirow{3}{*}{\textbf{V-H}}
    & Acc       & \textbf{0.795} & 0.700 & 0.860 & \textbf{0.810} & \textbf{0.878} & \textbf{0.903} & \textbf{0.849} \\
    & ROC-AUC   & \textbf{0.644} & \textbf{0.672} & \textbf{0.864} & \textbf{0.713} & \textbf{0.800} & \textbf{0.918} & \textbf{0.785} \\
    & MRA        & \textbf{0.829} & 0.743 & 0.905 & \textbf{0.858} & \textbf{0.891} & \textbf{0.949} & \textbf{0.857} \\
  \midrule
  \multirow{3}{*}{\textbf{H-H}}
    & Acc       & 0.747 & \textbf{0.722} & \textbf{0.911} & 0.737 & 0.659 & 0.885 & 0.692 \\
    & ROC-AUC   & 0.623 & 0.645 & 0.629 & 0.602 & 0.668 & 0.782 & 0.708 \\
    & MRA        & 0.807 & \textbf{0.757} & \textbf{0.919} & 0.820 & 0.825 & 0.868 & 0.824 \\
  \bottomrule
  \end{tabular}%
  }
  \vspace{2mm}
    \caption{Agreement between the automatic verifiers and the human annotators, benchmarked against inter-human agreement across three metrics. \textbf{Bold} indicates the higher value per column. The \textbf{V-H} row shows agreement between the verifiers and human annotators, and the \textbf{H-H} shows agreement within the human annotators.}
    \label{tab:verifier_agreement}
\end{table*}

%% file: tables/control_metrics.tex
\begin{table}[t]
\centering
\begin{tabular}{p{0.07\textwidth}l|cc}
\toprule
&Model & \makecell{Observ.} & \makecell{Action} \\
\midrule
 \multirow{5}{*}{\rotatebox{90}{\shortstack[l]{Video \\ models}}}
 & Veo 3          & \textbf{81.6} & \textbf{88.0} \\
 & Sora 2 Pro      & 69.4 & 80.4 \\
 & WAN 2.2        & 46.2 & 76.5  \\
 & HY-Video 1.5  & 31.2 & 81.0  \\
 & CogVideoX 1.5  & 22.2 & 75.5   \\
\midrule
\addlinespace[1pt]
 \multirow{5}{*}{\rotatebox{90}{\shortstack[l]{Camera-\\controlled}}}
& Genie 3          & 84.7 & \textbf{78.6}  \\
& HY-WorldPlay  & 55.4 & 64.9 \\
& Lingbot        & 35.6 &  67.8  \\
& GEN3C          & \textbf{90.6} & 57.6 \\
\addlinespace[3pt]
\bottomrule
\end{tabular}
\caption{Success rate of observation and action control. 
Models are largely able to follow control of \bench tasks.}
\label{table:controllability}
\end{table}


%% file: secs/06_conclusions.tex
\section{Conclusions}
We present \bench, a benchmark that evaluates whether video world models can evolve state correctly when observation is interrupted. We probe this by applying observation control, either as occlusions or illumination dimming for video models, or as ``lookaway'' camera commands for camera-controlled video models. \bench exposes that video world models struggle to evolve state when the observation control is applied. \bench's verifiers disentangle various failure modes to deepen our understanding of the limitations of today's video models. Evaluation on \bench yields interesting findings, such as video models' evolution stopping and incoherence failures, camera-controlled models' strong bias towards static scenes, and the increased difficulty in camera control when a dynamic evolution is present. These findings lead us to reflect on the training data and architecture design of video world models, such as training data bias the efficacy of memory-based architectures. \bench makes the first step toward evaluating the non-pixel-based ``world evolution'' aspect of world models. We hope \bench inspires new techniques and architectures that enable video world models to truly model processes in the world, and not just generate coherent pixels.

%% file: secs/07_acknowledgement.tex
\section{Acknowledgments}
We thank Aadarsh Sahoo and Damiano Marsili for valuable discussions. Ziqi Ma is funded by CAST. Mengzhan Liufu is supported by the Cherng fellowship. Georgia acknowledges the William Hurt Scholarship program for their support. We also thank Google for generously providing us with Gemini credits.

%% file: secs/X_supp.tex
\clearpage
\appendix
\setcounter{page}{1}
\maketitlesupplementary

\section{Benchmark}

\begin{figure*}[h!]
    \centering
    \includegraphics[width=\linewidth]{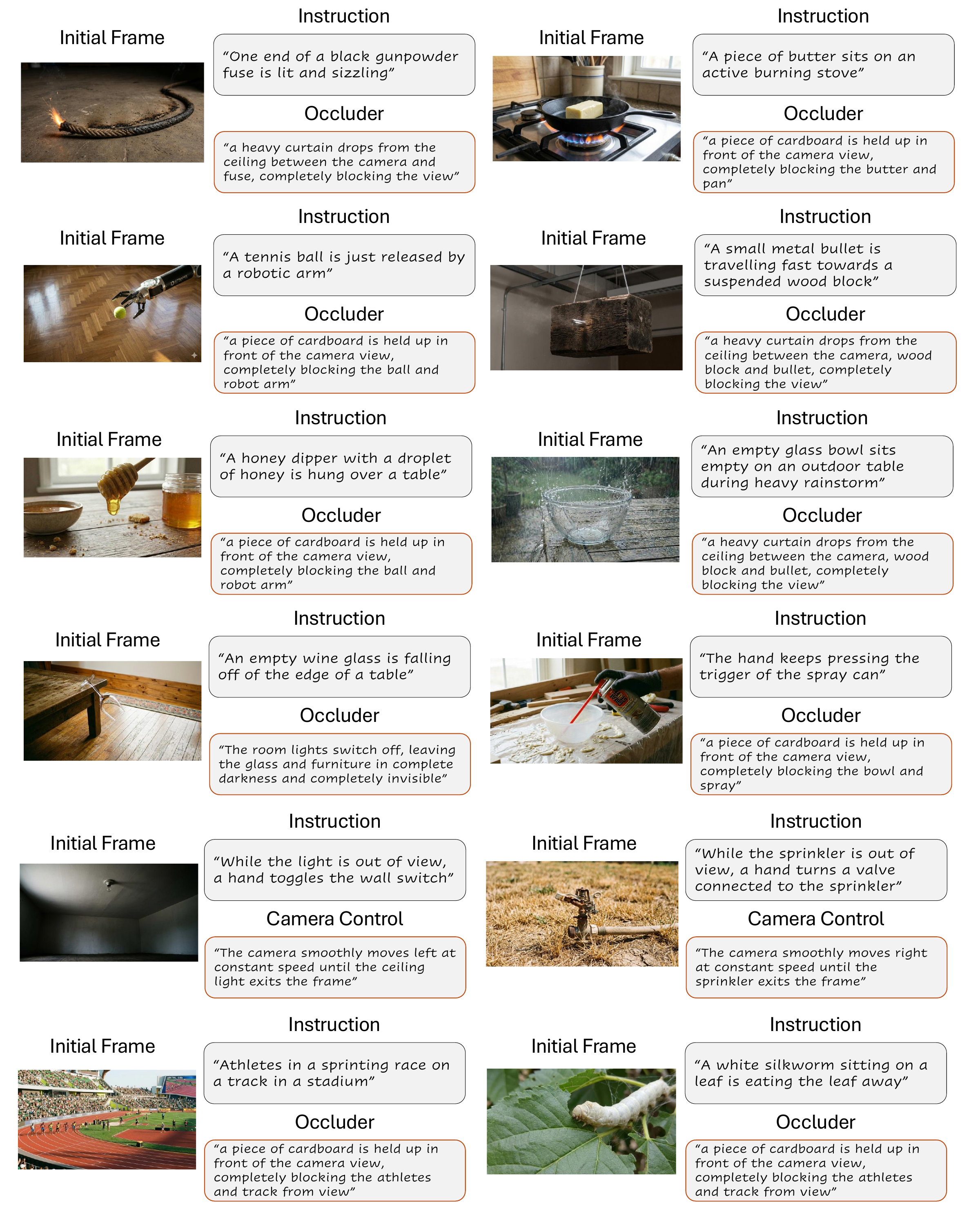}
    \caption{Example tasks from \bench. Each row shows two tasks from a unique category. Category from top to bottom: continuous evolution, kinematics, relational change, state transformation, causal change, and expected human \& animal behavior.}
    \label{fig:benchmark_examples}
\end{figure*}

We present additional tasks of \bench in Fig. \ref{fig:benchmark_examples}. These example tasks cover all six task categories in \bench, each of which captures a fundamentally distinct category of state evolution. The categories span continuous evolutions that accumulate gradually over time, simple few-object kinematics, irreversible structural or material transformations, causal change, and expected behavior of humans or animals. Together, they cover the physical, chemical, and social dimensions of world dynamics, providing a comprehensive probe of a model's internal simulation capacity. Coupled with our diverse methods of observation control, strong performance on \bench reflects genuine, real-world-relevant understanding. 

\section{Verifiers}
\subsection{Qualitative Examples}
\begin{figure*}[h!]
    \centering
    \includegraphics[width=\linewidth]{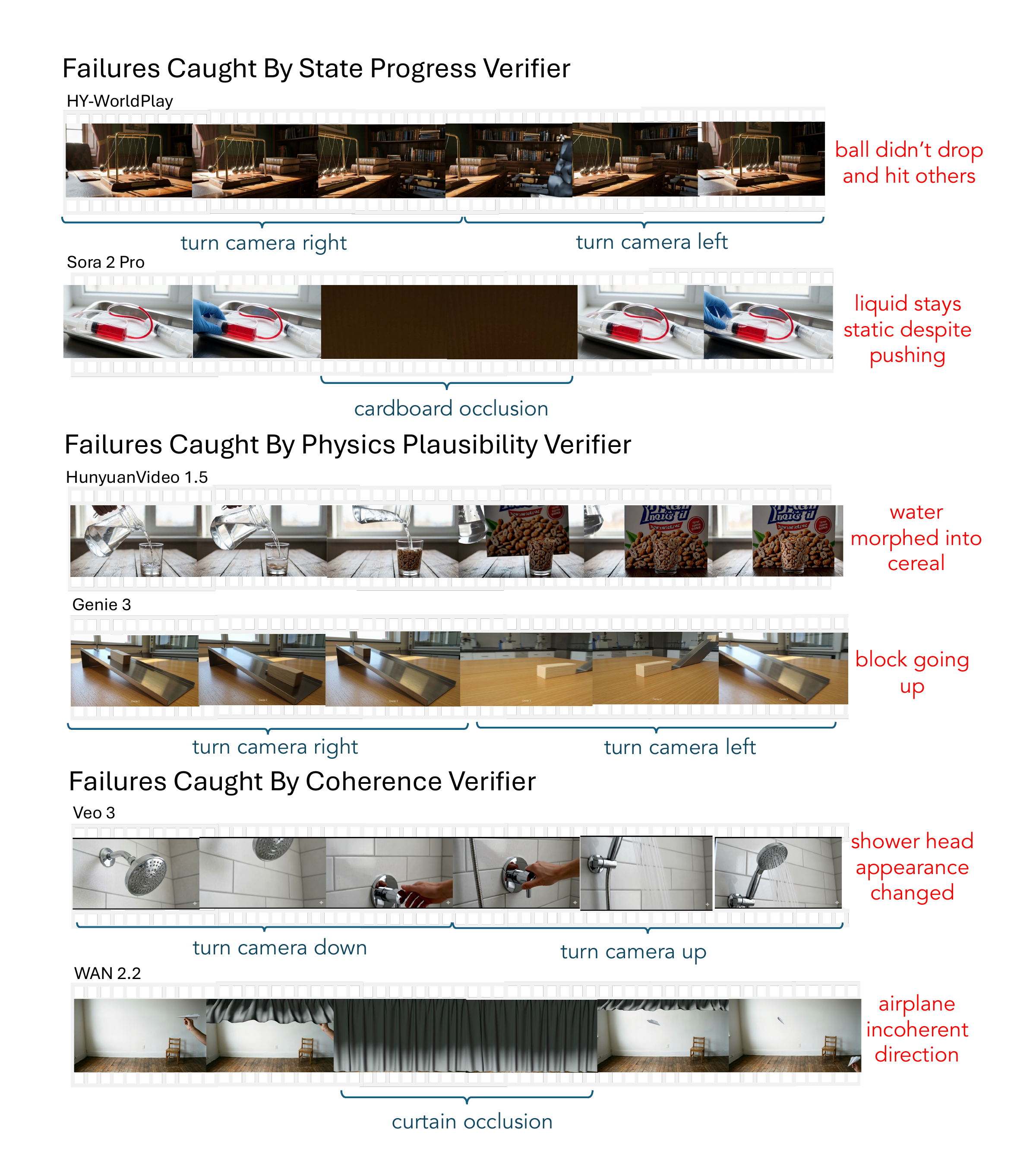}

    \caption{Example video generation failures that violate verifiers on the ``state evolution'' axes. We show separate examples that violate the state progress verifier, physical plausibility verifier, and the coherence verifier.}
    \label{fig:verifier_failure_evo}
\end{figure*}

\begin{figure*}[h!]
    \centering
    \includegraphics[width=\linewidth]{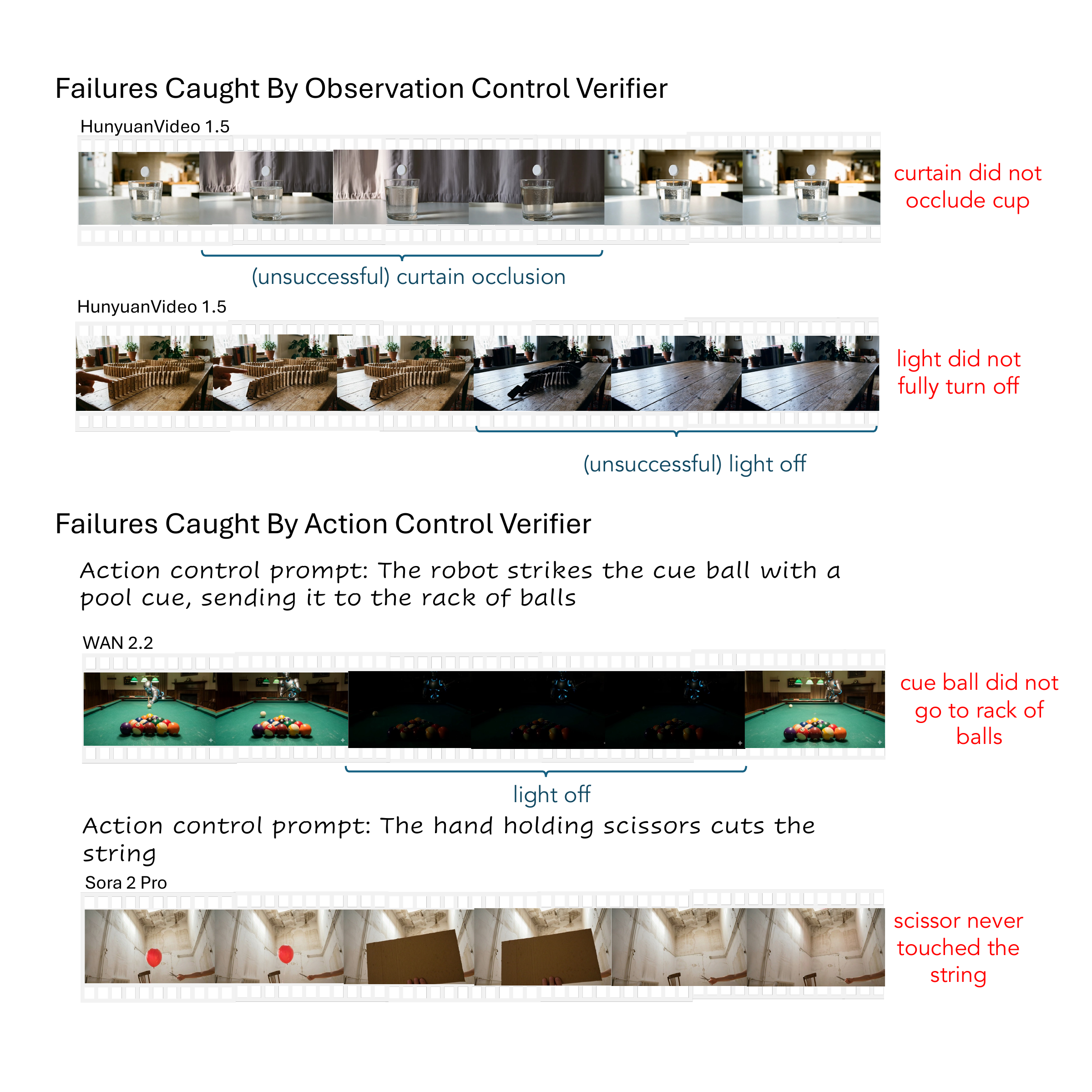}

    \caption{Example video generation failures that violate verifiers on the ``control'' axes. We show separate examples that violate the observation control verifier and action control verifier.}
    \label{fig:verifier_failure_control}
\end{figure*}

We provide additional examples of video failures that are caught by each verifier in \cref{fig:verifier_failure_evo} and \cref{fig:verifier_failure_control}. \cref{fig:verifier_failure_evo} provides examples of videos that fail to evolve state correctly, either due to failure in state progress, physical implausibility, or incoherence. \cref{fig:verifier_failure_control} shows examples of videos that fail to adhere to specified control, either failing to insert the correct observation control or failing to perform the action that initiates the state evolution.

\subsection{Example Verifier Rationale}
\begin{figure*}[h!]
    \centering
    \includegraphics[width=\linewidth]{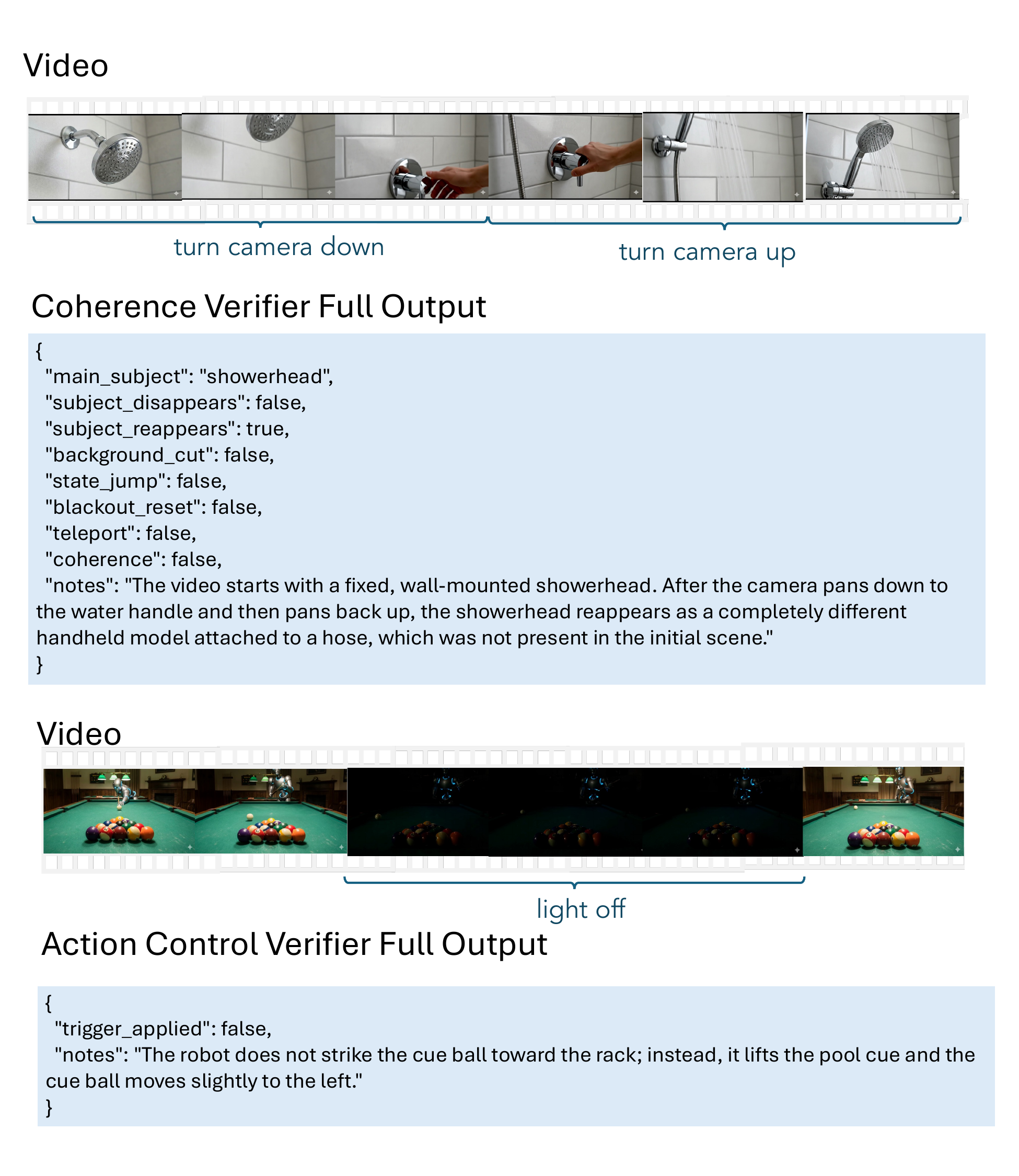}
    \caption{Example verifier rationales. The verifier evaluates various aspects defined for each evaluation axis (when applicable), and provides an explanation of how the video violates the corresponding criteria.}
    \label{fig:verifier_rationale}
\end{figure*}

\cref{fig:verifier_rationale} shows two examples of the verifier rationale. The verifier considers various aspects of the criteria, and provides a short analysis on the violating aspect.

\subsection{Further Verifier-Human Agreement Analysis}

\begin{figure*}[h!]
    \centering
    \includegraphics[width=0.6\textwidth]{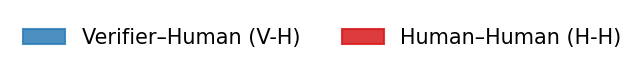}
    \vspace{-1.5mm} 
    \includegraphics[width=\textwidth]{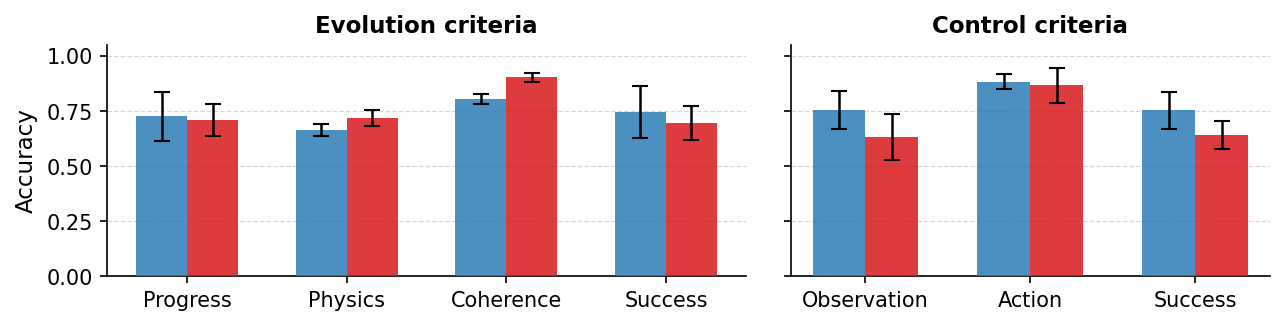}
    \vspace{-1.5mm} 
    \includegraphics[width=\textwidth]{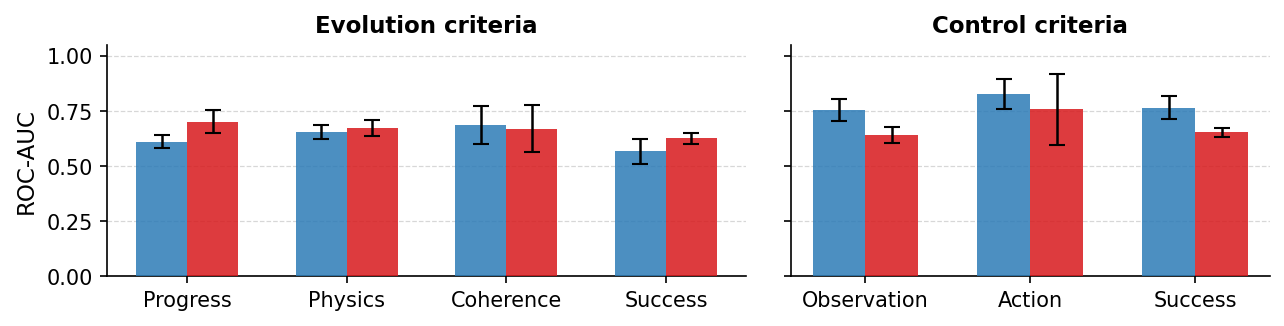}
    \vspace{-1.5mm} 
    \includegraphics[width=\textwidth]{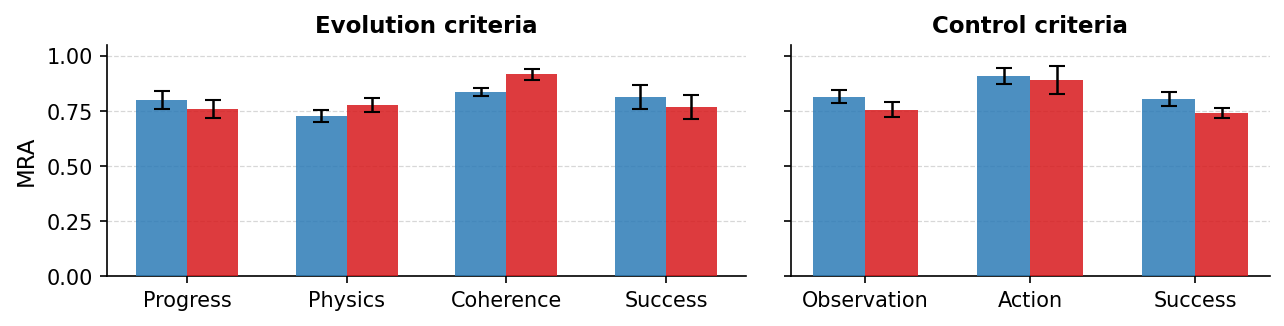}

    \caption{Mean agreement and standard deviation across three Verifier-Human (V-H) pairs and three Human-Human (H-H) pairs for each criterion. Agreement measured by \texttt{Accuracy}, \texttt{ROC-AUC} and \texttt{Model Ranking Agreement (MRA)} is shown in top, middle and bottom panel respectively.}
    \label{fig:agreement_bar}
\end{figure*}

\begin{figure*}[h!]
    \centering
    \begin{minipage}{0.8\textwidth} 
        \centering
        
        \newcommand{\smallplot}[1]{%
            \includegraphics[width=0.48\linewidth, height=0.48\linewidth, keepaspectratio]{#1}%
        }

        \smallplot{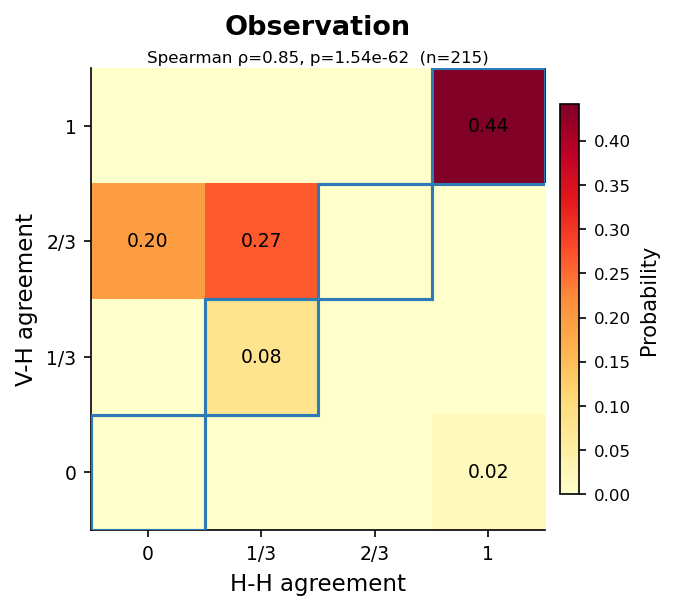}\hfill
        \smallplot{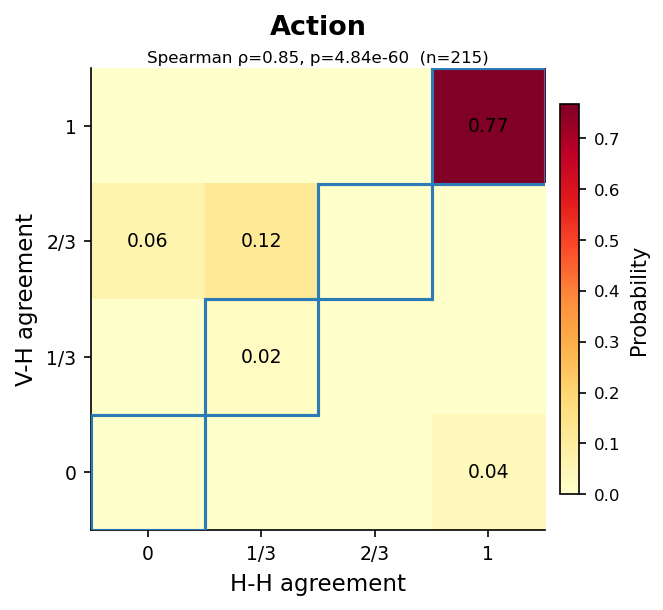}
        
        \vspace{-0.5mm} 
        
        \smallplot{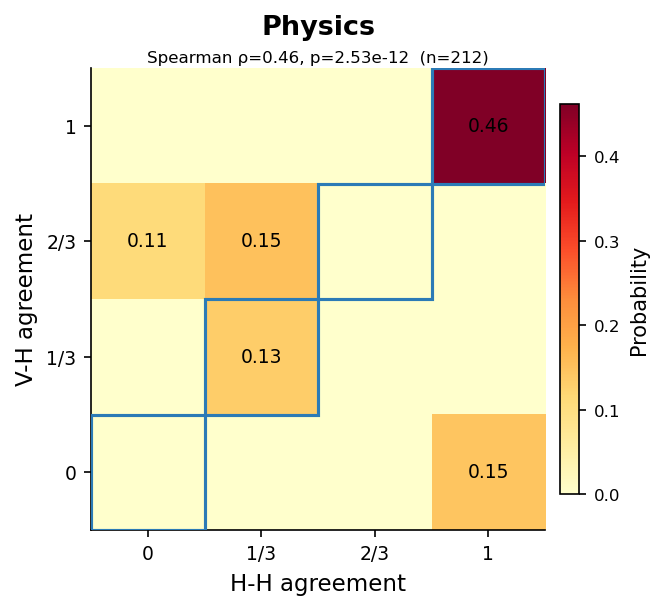}\hfill
        \smallplot{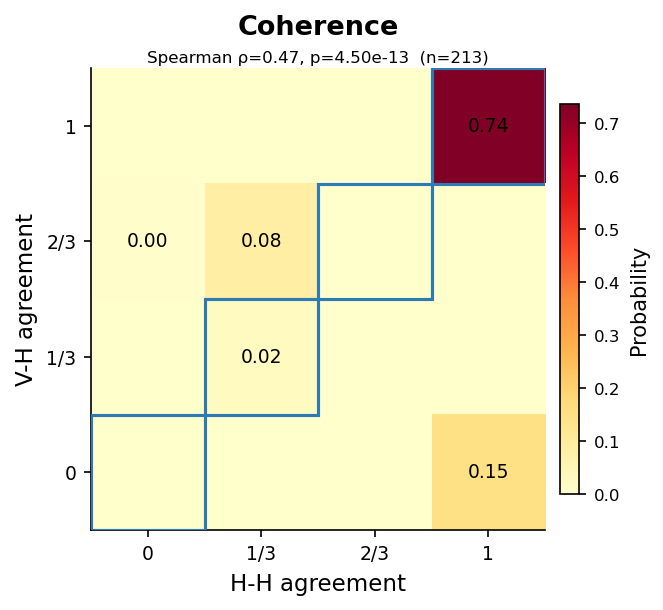}
        
        \vspace{-0.5mm} 
        
        \smallplot{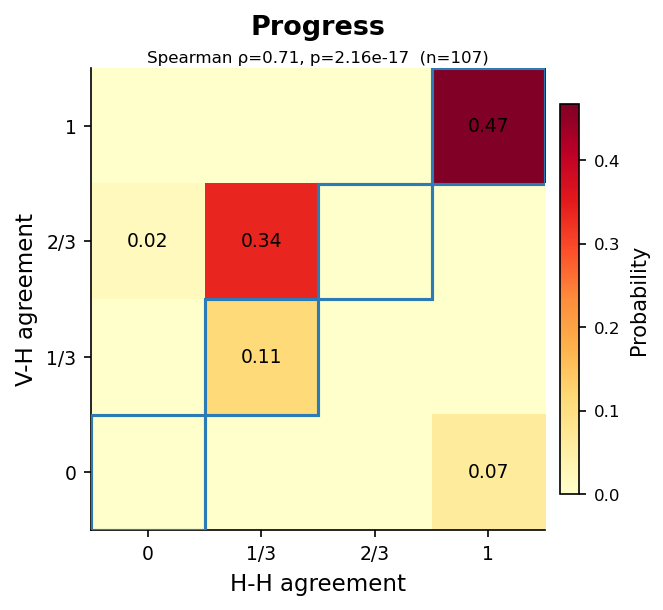}
    \end{minipage}
    \caption{Agreement heatmap for each criterion. V-H agreement and H-H agreement are calculated using Accuracy. With 3 human annotators, both V-H and H-H agreement can only take discrete values from the set $\{0, 1/3, 2/3, 1\}$. Correlation between V-H agreement and H-H agreement is calculated using Spearman's correlation coefficient.}
    \label{fig:agreement_corr}
\end{figure*}

We present further analysis on agreement between the automatic verifiers and the human annotators. Mean agreement measured using three metrics and their respective standard deviation are shown in Fig. \ref{fig:agreement_bar}. The standard deviation is low across all criteria and agreement metrics, meaning the behavior of both the verifier and the human annotators are consistent across all comparison pairs. This consistency assures that the Verifier-Human (V-H) agreement and Human-Human (H-H) agreement measures are genuine and reliable. We also found strong correlation between V-H agreement and H-H agreement for every criterion across tasks (Fig. \ref{fig:agreement_corr}). The correlation coefficient has $p < 10^{-11}$ for all criteria. This means if the automatic verifier disagrees with the human annotators on a video, the video is likely also hard for human annotators to agree on. This confirms that the automatic verifiers agree with human annotators at least as well as human annotators agree with each other. 

\subsection{Disagreements}

\begin{figure*}[h!]
    \centering
    \includegraphics[width=\linewidth]{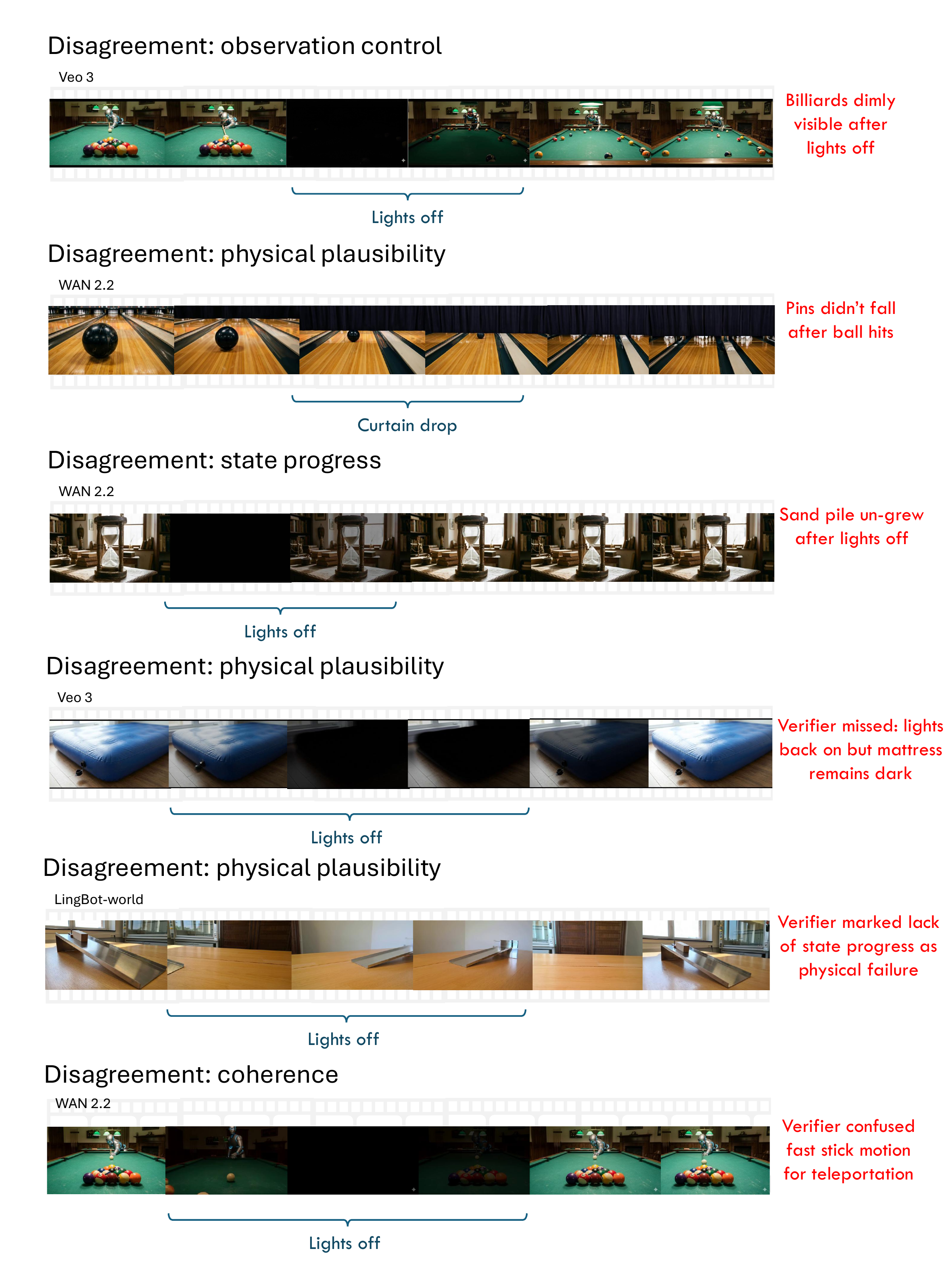}

    \caption{Example videos where the verifier and human annotators disagree with each other. The top 3 disagreements are caused by ambiguity in the generated videos, and bottom 3 disagreements show verifier failures when they overlooked aspects of the videos or failed to follow the prompts. }
    \label{fig:disagreement_example}
\end{figure*}

We analyze videos where the verifiers and human annotators disagree. Example disagreements are shown in Fig. \ref{fig:disagreement_example}. These often coincide with videos where human annotators disagree with each other. 

Most videos where the verifiers and human annotators disagree exhibit some inherent ambiguity. The ambiguity stems from at least two distinct sources. First, sometimes a quantitative thresholding judgement is required: for example, how much deflation constitutes a fully deflated mattress? how dark does the room have to be for the ``lights off'' control to be considered successful? The verifiers and human annotators might apply this threshold differently. Second, each party must decide which visual discrepancies are consequential enough to fail a video. A minor physics inconsistency or a subtle continuity glitch may be dismissed as irrelevant noise by one annotator yet treated as a meaningful failure by another. This reflects a subjective salience judgment about what the criterion is ultimately testing. Together, these ambiguity make definitive binary labeling difficult for some videos. When such ambiguity is present, the observed V-H and H-H disagreement should be interpreted as a property of the evaluation task itself rather than a limitation of the verifiers or the human annotators.

Occasionally, verifier errors do occur from VLM failures. The VLM may not follow the verifier prompt instructions precisely, or may overlook a subtle but decisive visual detail (Fig. \ref{fig:disagreement_example}). While these reflect occasional, stochastic failures of the underlying VLM, we mitigate them by the ensemble design of our verifiers. By querying the VLM multiple times per judgment (with rephrased prompts, when applicable) and aggregating votes, we prevent single lapses from contaminating the final judgments in many tasks.

\clearpage

\subsection{Verifier Prompts}

The full prompt of each automatic verifier is provided below.

\input{secs/X01_verifier_prompts}

%% file: secs/X01_verifier_prompts.tex
\begin{strip}
\begin{tcolorbox}[
    colback=boxblue,
    colframe=blue!50!black,
    arc=0mm,
    sharp corners,
    fontupper=\rmfamily\verifierfontsize, 
    title=Observation Control Verifier Prompt,
    fonttitle=\bfseries\rmfamily,
    left=5pt,
    right=5pt,
    breakable
]
\begin{lstlisting}[basicstyle=\rmfamily\verifierfontsize, breaklines=True, breakindent=0pt, columns=fullflexible]
You are evaluating a generated video for whether a partial observability mechanism was correctly applied.

The following has been determined from the task specification:
requested_occlusion: {occlusion_display}

PARTIAL OBSERVABILITY:
The video may hide the scene via in-scene occlusion (lights off, object placed in front, smoke filling the view, etc.) OR by camera pan (camera moves away, taking the subject fully out of frame). Both mechanisms are valid.

Do NOT penalise a mismatch between requested_occlusion and the mechanism shown in the video. Occlusion success is judged solely by whether the subject became invisible, regardless of how.

Evaluate ONE thing from the GENERATED video: occlusion_done - Were the key subject and dynamic successfully hidden from view (became invisible) at the appropriate moment, by ANY mechanism?

Camera pan that moves the subject fully out of frame counts as occlusion_done = True, even if requested_occlusion described a different in-scene method - and vice versa.

TRUE only if the main subject/area became clearly invisible or fully obscured.
FALSE if the dynamic process already completed BEFORE the occlusion took place.
FALSE if the scene was never hidden at all.
Set to TRUE if requested_occlusion is "none".

General guidance:
- Use only visual evidence from the video.
- Ignore timestamps, watermarks, subtitles, and UI overlays.

Return ONLY valid JSON in this exact format (no markdown, no commentary):
{
  "occlusion_done": true/false,
  "notes": "optional short explanation"
}
\end{lstlisting}
\end{tcolorbox}

\begin{tcolorbox}[
    breakable,
    colback=boxgreen, 
    colframe=green!40!black, 
    title=Action Control Verifier Prompt, 
    fonttitle=\bfseries\rmfamily, 
    sharp corners, 
    fontupper=\rmfamily\verifierfontsize,
    left=5pt,
    right=5pt
]
\begin{lstlisting}[
    basicstyle=\rmfamily\verifierfontsize, 
    breaklines=true, 
    breakindent=0pt, 
    columns=fullflexible
]
"You are evaluating a generated video for whether a trigger action was correctly applied.

The following has been determined from the task specification:
requested_trigger: {trigger_display}

Evaluate ONE thing from the GENERATED video:

trigger_applied - Did the requested kickoff action occur?
TRUE if the action itself happened or its physical effect is clearly visible.
Set to true if requested_trigger is "none".

General guidance:
- Use only visual evidence from the video.
- Ignore timestamps, watermarks, subtitles, and UI overlays.

Return ONLY valid JSON in this exact format (no markdown, no commentary):
{
  "trigger_applied": true/false,
  "notes": "optional short explanation"
}
\end{lstlisting}
\end{tcolorbox}

\begin{tcolorbox}[
    breakable,
    colback=boxpurple, 
    colframe=purple!50!black, 
    title=State Progress Verifier Prompt, 
    fonttitle=\bfseries\rmfamily, 
    sharp corners,  
    fontupper=\rmfamily\verifierfontsize,
    left=5pt,
    right=5pt
]
\begin{lstlisting}[
    basicstyle=\rmfamily\verifierfontsize, 
    breaklines=true, 
    breakindent=0pt, 
    columns=fullflexible
]
Your job is to verify whether a video shows an expected physical process.

The expected process is:
1. An initial image showing the starting state
2. An action/event that will occur: "{action_prompt}"

Based on the initial image and the action described, what physical process or state evolution should happen?
Describe the expected changes in the scene, including:
- What objects or elements will change
- How they will change (e.g., shape, position, state)
- What the final state should look like

IMPORTANT: This video includes CAMERA MOVEMENT where the camera turns away from the main object.
You must specifically check: Did the expected process HAPPEN while the camera was turned away (when the main object was not visible)?
This is critical - the process should evolve even when the camera is not looking at it.

Watch the video carefully and answer:
1. Did the expected process happen in the video? (Yes/No)
2. Provide a detailed explanation of what is the object state right before it becomes invisible, and right after it is seen again.
3. Compare what happened in the video to what was expected
4. SPECIFICALLY address: Did the process continue/happen during the occlusion or camera movement?
5. If the process did not happen as expected, explain what was different or missing

Please ignore physical violations or artifacts. Only focus on whether the key process happened DURING the occlusion or camera movement.

After your full explanation, provide a one-sentence summary on the last line in this format:
Process: [concise one-sentence description of the expected evolution]

Format your response as:
VERDICT: [Yes/No]
EXPLANATION: [Your detailed explanation]
\end{lstlisting}
\end{tcolorbox}

\begin{tcolorbox}[
    colback=boxrose, 
    colframe=red!50!black, 
    title=Prompt 3: Physical Plausibility Verifier Prompt, 
    fonttitle=\bfseries\rmfamily, 
    sharp corners, 
    fontupper=\rmfamily\verifierfontsize,
    left=5pt,
    right=5pt,
]
\begin{lstlisting}[
    basicstyle=\rmfamily\verifierfontsize, 
    breaklines=true, 
    breakindent=0pt, 
    columns=fullflexible
]
You are evaluating a generated video for physical plausibility.

An ARTIFACT is any event or behavior in the video that actively violates the laws of physics or is otherwise physically impossible. There are two types:

TYPE 1 - Instantaneous violations
Something is physically wrong in a single frame or very short moment, independent of what came before or after.
Examples:
- A rigid object (cup, block, bottle) deforms or changes shape without any force being applied
- An object's color or texture changes suddenly without a physical cause
- Two solid objects overlap, pass through each other, or merge together
- An object abruptly moves or changes velocity with no visible contact, force, or physical cause
- An object floats in mid-air against gravity, without support or attachment

TYPE 2 - Dynamic violations
Individual frames look plausible, but the evolution of the scene over time violates physics. The cause is shown, but the WRONG effect follows.
Examples:
- Water is continuously poured into a glass but the level drops or stays flat (the wrong direction of change given the cause).
- A block slides down a ramp but then reverses upward with no visible push.
- An object is set in motion but accelerates with no driving force.

For TYPE 2, ask yourself: a cause is clearly shown - does the effect that follows actively defy physics? If yes, that is an artifact. If instead the scene simply does not change much (the cause has no visible effect), that is a state evolution failure, not an artifact - do NOT flag it here.

INTENDED OCCLUSION CONTEXT: This task intentionally includes the following occlusion: "{requested_occlusion}". Any scene darkening, blackout, or view obstruction that corresponds to this description is expected behaviour and must NOT be flagged as an artifact. Do NOT flag any irregularities related to the intended occlusion as artifacts. If the intended occlusion does not correctly happen, DO NOT flag it as artifact either.

INTENDED STATE EVOLUTION: The following image-edit prompt describes the main physical change that was intended to occur in the scene:
"{init2final_edit_prompt}"

Before judging artifacts, extract a one-sentence plain-language summary of this intended change (e.g. "the block slides to the bottom of the ramp") and record it as "intended_state_evolution" in your JSON response.

CRITICAL DISTINCTION - artifact vs. state evolution failure:
- ARTIFACT: something WRONG actively happens (a physically impossible event occurs, e.g. an object deforms with no force, two solids pass through each other, water level drops while being poured in).
- STATE EVOLUTION FAILURE: the intended change simply does NOT happen (e.g. the scene looks frozen, nothing moves, the intended effect is absent). This is evaluated separately and must NOT be flagged here.

Only flag as artifact if the video shows an active physical violation, not merely a lack of the intended change.

SEVERITY THRESHOLD: Only flag violations that a casual viewer would find clearly jarring and physically impossible. Do NOT flag:
- Subtle rendering imperfections (slightly uneven textures, minor flickering)
- Small quantitative discrepancies (smoke that diffuses slightly faster than expected)
- Sand or granular material appearing or clumping in plausible ways
- Any effect that, while imperfect, is physically plausible
- Video quality issues such as blurriness or noise

General guidance:
- Watch the full video before making a judgment.
- Flag only CLEAR, VISUALLY JARRING violations that a layperson would immediately notice as impossible.
- Use only visual evidence. Ignore timestamps, watermarks, and UI overlays.

Return ONLY valid JSON in this exact format (no markdown, no commentary):
{
  "intended_state_evolution": "one-sentence summary of the intended physical change",
  "artifact": true/false,
  "notes": "brief description of what was observed and why it is or is not an artifact"
}
\end{lstlisting}
\end{tcolorbox}

\begin{tcolorbox}[
    colback=boxblue, 
    colframe=blue!50!black, 
    title=Prompt 4: Coherence Verifier Prompt, 
    fonttitle=\bfseries\rmfamily\verifierfontsize, 
    sharp corners, 
    breakable, 
    fontupper=\rmfamily\verifierfontsize,
    left=10pt,
    right=10pt,
]
\begin{lstlisting}[
    basicstyle=\rmfamily\verifierfontsize, 
    breaklines=true, 
    breakindent=0pt, 
    columns=fullflexible
]
You are evaluating a generated video for temporal and scene coherence.

A COHERENCE FAILURE means the video does not appear to be a single, continuous, uninterrupted recording of one scene.

INTENDED OCCLUSION: This task uses a CAMERA PAN to hide the subject.
The camera sweeps the subject out of frame, then pans back to reveal it.
This is intentional and expected behaviour.

OCCLUSION vs. DISAPPEARANCE for camera pan:
- Subject leaving the frame because the camera panned away = OCCLUSION. Mark subject_disappears = false in this case.
- Only mark subject_disappears = true if the subject vanishes while still within the camera's field of view, with no physical cause.

Camera pan vs. scene cut:
A pan has continuous intermediate frames of spatial motion connecting the two viewpoints. A cut jumps INSTANTANEOUSLY with zero transitional frames. If any sliding motion is visible, it is a pan - not a cut. Do NOT mark background_cut = true for a camera pan, even a fast or direction-reversing one.

STEP 1 - Identify the main subject.
Identify the primary object or entity the video is about. Record it as main_subject.

STEP 2 - Occlusion vs. disappearance.
For this task, determine whether any disappearance of the main subject is explained by the INTENDED OCCLUSION described above.
- OCCLUSION: subject becomes invisible because something hides it (camera moves away, lights off, curtain drawn, smoke fills frame, etc.). The subject is still physically present - just hidden. NOT a failure.
- DISAPPEARANCE: subject is simply absent with no physical or occlusion cause. This IS a coherence failure.

STEP 3 - Fill in the checklist.
Answer each item true or false based only on visual evidence:

subject_disappears - Did the main subject vanish from the scene with NO physical explanation and NO correspondence to the intended occlusion? (If the intended occlusion accounts for the disappearance, mark false.)

subject_reappears - After the subject was hidden by occlusion, did it reappear in a state or position that cannot be explained by the visible process? (A normal reveal after intended occlusion is fine - flag only unexpected discontinuous jumps upon reappearance.)

background_cut - Did the background or environment change INSTANTANEOUSLY (zero transitional frames), implying a hidden edit or scene cut? Example: table surface or wall color abruptly changes; a window switches sides. A camera pan has continuous motion - it is NOT a cut.

state_jump - Did the main subject's state change discontinuously, with no visible transition? Example: a full glass is suddenly empty; an intact ice cube is suddenly fully melted, with the melting process entirely skipped.

blackout_reset - Did the scene briefly black out and then resume in a visibly different configuration (objects repositioned or in different states) with no physical explanation? (A blackout matching the intended occlusion that resumes the same scene is NOT a reset.)

teleport - Did any object's position jump discontinuously between frames, with no continuous motion connecting the two locations? Example: a block on the left of a ramp is suddenly at the bottom right.

COHERENCE RULE:
coherence = false if ANY checklist item above is true.
coherence = true if ALL checklist items are false.

Do NOT flag:
- Normal continuous physical processes, even surprising or fast ones.
- Video quality issues such as blurriness or noise.
- subject_disappears = false when the subject leaves frame due to camera pan.
- background_cut = false for any camera pan motion, including fast or direction-reversing pans (pan away to hide, pan back to reveal).

General guidance:
- Watch the full video before filling the checklist.
- Flag only CLEAR failures. Unusual but continuous events are not failures.
- Use only visual evidence. Ignore timestamps, watermarks, and UI overlays.

Return ONLY valid JSON in this exact format (no markdown, no commentary):
{
  "main_subject": "<primary object or entity in the video>",
  "subject_disappears": true/false,
  "subject_reappears": true/false,
  "background_cut": true/false,
  "state_jump": true/false,
  "blackout_reset": true/false,
  "teleport": true/false,
  "coherence": true/false,
  "notes": "brief explanation referencing specific observations"
}
(coherence: true = video IS coherent - all checklist items false; coherence: false = a coherence failure was detected.)
\end{lstlisting}
\end{tcolorbox}
\end{strip}

%% file: main.bib
@String(ICCV= {Int. Conf. Comput. Vis.})

@String(ICCV  = {ICCV})

@String(ICCV  = {Int. Conf. Comput. Vis.})

@String(NeurIPS = {Adv. Neural Inform. Process. Syst.})

@String(NeurIPS = {NeurIPS})

@InProceedings{duan2025worldscore,
    author    = {Duan, Haoyi and Yu, Hong-Xing and Chen, Sirui and Fei-Fei, Li and Wu, Jiajun},
    title     = {WorldScore: A Unified Evaluation Benchmark for World Generation},
    booktitle = {Proceedings of the IEEE/CVF International Conference on Computer Vision (ICCV)},
    month     = {October},
    year      = {2025},
    pages     = {27713-27724}
}

@InProceedings{huang2023vbench,
    title={{VBench}: Comprehensive Benchmark Suite for Video Generative Models},
    author={Huang, Ziqi and He, Yinan and Yu, Jiashuo and Zhang, Fan and Si, Chenyang and Jiang, Yuming and Zhang, Yuanhan and Wu, Tianxing and Jin, Qingyang and Chanpaisit, Nattapol and Wang, Yaohui and Chen, Xinyuan and Wang, Limin and Lin, Dahua and Qiao, Yu and Liu, Ziwei},
    booktitle={Proceedings of the IEEE/CVF Conference on Computer Vision and Pattern Recognition},
    year={2024}
}

@article{zheng2025vbench2,
    title={{VBench-2.0}: Advancing Video Generation Benchmark Suite for Intrinsic Faithfulness},
    author={Zheng, Dian and Huang, Ziqi and Liu, Hongbo and Zou, Kai and He, Yinan and Zhang, Fan and Zhang, Yuanhan and He, Jingwen and Zheng, Wei-Shi and Qiao, Yu and Liu, Ziwei},
    journal={arXiv preprint arXiv:2503.21755},
    year={2025}
}

@misc{mak2026physicsmindsimrealmechanics,
      title={PhysicsMind: Sim and Real Mechanics Benchmarking for Physical Reasoning and Prediction in Foundational VLMs and World Models}, 
      author={Chak-Wing Mak and Guanyu Zhu and Boyi Zhang and Hongji Li and Xiaowei Chi and Kevin Zhang and Yichen Wu and Yangfan He and Chun-Kai Fan and Wentao Lu and Kuangzhi Ge and Xinyu Fang and Hongyang He and Kuan Lu and Tianxiang Xu and Li Zhang and Yongxin Ni and Youhua Li and Shanghang Zhang},
      year={2026},
      eprint={2601.16007},
      archivePrefix={arXiv},
      primaryClass={cs.CV},
      url={https://arxiv.org/abs/2601.16007}, 
}

@misc{wu2026considgenviewconsistentidentitypreservingimagetovideo,
      title={ConsID-Gen: View-Consistent and Identity-Preserving Image-to-Video Generation}, 
      author={Mingyang Wu and Ashirbad Mishra and Soumik Dey and Shuo Xing and Naveen Ravipati and Hansi Wu and Binbin Li and Zhengzhong Tu},
      year={2026},
      eprint={2602.10113},
      archivePrefix={arXiv},
      primaryClass={cs.CV},
      url={https://arxiv.org/abs/2602.10113}, 
}

@misc{ye2026mindbenchmarkingmemoryconsistency,
      title={MIND: Benchmarking Memory Consistency and Action Control in World Models}, 
      author={Yixuan Ye and Xuanyu Lu and Yuxin Jiang and Yuchao Gu and Rui Zhao and Qiwei Liang and Jiachun Pan and Fengda Zhang and Weijia Wu and Alex Jinpeng Wang},
      year={2026},
      eprint={2602.08025},
      archivePrefix={arXiv},
      primaryClass={cs.CV},
      url={https://arxiv.org/abs/2602.08025}, 
}

@misc{veo2025,
  title        = {Veo: a text-to-video generation system},
  author       = {{Google DeepMind}},
  year         = {2025},
  howpublished = {\url{https://storage.googleapis.com/deepmind-media/veo/Veo-3-Tech-Report.pdf}},
  note         = {Accessed: 2026-03-03}
}

@misc{genie2025,
  title        = {Genie 3: A new frontier for world models},
  author       = {{Google DeepMind}},
  year         = {2025},
  howpublished = {\url{https://deepmind.google/blog/genie-3-a-new-frontier-for-world-models/}},
  note         = {Accessed: 2026-03-03}
}

@misc{sora2025,
  title        = {Sora 2},
  author       = {{OpenAI}},
  year         = {2025},
  howpublished = {\url{https://openai.com/index/sora-2/}},
  note         = {Accessed: 2026-03-03}
}

@article{wan2025,
      title={Wan: Open and Advanced Large-Scale Video Generative Models}, 
      author={Team Wan and Ang Wang and Baole Ai and Bin Wen and Chaojie Mao and Chen-Wei Xie and Di Chen and Feiwu Yu and Haiming Zhao and Jianxiao Yang and Jianyuan Zeng and Jiayu Wang and Jingfeng Zhang and Jingren Zhou and Jinkai Wang and Jixuan Chen and Kai Zhu and Kang Zhao and Keyu Yan and Lianghua Huang and Mengyang Feng and Ningyi Zhang and Pandeng Li and Pingyu Wu and Ruihang Chu and Ruili Feng and Shiwei Zhang and Siyang Sun and Tao Fang and Tianxing Wang and Tianyi Gui and Tingyu Weng and Tong Shen and Wei Lin and Wei Wang and Wei Wang and Wenmeng Zhou and Wente Wang and Wenting Shen and Wenyuan Yu and Xianzhong Shi and Xiaoming Huang and Xin Xu and Yan Kou and Yangyu Lv and Yifei Li and Yijing Liu and Yiming Wang and Yingya Zhang and Yitong Huang and Yong Li and You Wu and Yu Liu and Yulin Pan and Yun Zheng and Yuntao Hong and Yupeng Shi and Yutong Feng and Zeyinzi Jiang and Zhen Han and Zhi-Fan Wu and Ziyu Liu},
      journal = {arXiv preprint arXiv:2503.20314},
      year={2025}
}

@article{kong2024hunyuanvideo,
  title={Hunyuanvideo: A systematic framework for large video generative models},
  author={Kong, Weijie and Tian, Qi and Zhang, Zijian and Min, Rox and Dai, Zuozhuo and Zhou, Jin and Xiong, Jiangfeng and Li, Xin and Wu, Bo and Zhang, Jianwei and others},
  journal={arXiv preprint arXiv:2412.03603},
  year={2024}
}

@misc{zheng2026opensora20trainingcommerciallevel,
      title={Open-Sora 2.0: Training a Commercial-Level Video Generation Model in \$200k}, 
      author={Zangwei Zheng and Xiangyu Peng and Yuxuan Lou and Chenhui Shen and Tom Young and Xinying Guo and Binluo Wang and Hang Xu and Hongxin Liu and Mingyan Jiang and Wenjun Li and Yuhui Wang and Anbang Ye and Gang Ren and Qianran Ma and Wanying Liang and Xiang Lian and Xiwen Wu and Yuting Zhong and Zhuangyan Li and Chaoyu Gong and Guojun Lei and Leijun Cheng and Limin Zhang and Minghao Li and Ruijie Zhang and Silan Hu and Shijie Huang and Xiaokang Wang and Yuanheng Zhao and Yuqi Wang and Ziang Wei and Yang You},
      year={2026},
      eprint={2503.09642},
      archivePrefix={arXiv},
      primaryClass={cs.GR},
      url={https://arxiv.org/abs/2503.09642}, 
}

@article{sun2025worldplay,
  title={Worldplay: Towards long-term geometric consistency for real-time interactive world modeling},
  author={Sun, Wenqiang and Zhang, Haiyu and Wang, Haoyuan and Wu, Junta and Wang, Zehan and Wang, Zhenwei and Wang, Yunhong and Zhang, Jun and Wang, Tengfei and Guo, Chunchao},
  journal={arXiv preprint arXiv:2512.14614},
  year={2025}
}

@article{lingbot-world,
      title={Advancing Open-source World Models}, 
      author={Robbyant Team and Zelin Gao and Qiuyu Wang and Yanhong Zeng and Jiapeng Zhu and Ka Leong Cheng and Yixuan Li and Hanlin Wang and Yinghao Xu and Shuailei Ma and Yihang Chen and Jie Liu and Yansong Cheng and Yao Yao and Jiayi Zhu and Yihao Meng and Kecheng Zheng and Qingyan Bai and Jingye Chen and Zehong Shen and Yue Yu and Xing Zhu and Yujun Shen and Hao Ouyang},
      journal={arXiv preprint arXiv:2601.20540},
      year={2026}
}

@inproceedings{bar2025navigation,
  title={Navigation world models},
  author={Bar, Amir and Zhou, Gaoyue and Tran, Danny and Darrell, Trevor and LeCun, Yann},
  booktitle={Proceedings of the Computer Vision and Pattern Recognition Conference},
  pages={15791--15801},
  year={2025}
}

@article{ali2025world,
  title={World simulation with video foundation models for physical ai},
  author={Ali, Arslan and Bai, Junjie and Bala, Maciej and Balaji, Yogesh and Blakeman, Aaron and Cai, Tiffany and Cao, Jiaxin and Cao, Tianshi and Cha, Elizabeth and Chao, Yu-Wei and others},
  journal={arXiv preprint arXiv:2511.00062},
  year={2025}
}

@article{wang2025bullettime,
  title={BulletTime: Decoupled Control of Time and Camera Pose for Video Generation},
  author={Wang, Yiming and Zhang, Qihang and Cai, Shengqu and Wu, Tong and Ackermann, Jan and Kuang, Zhengfei and Zheng, Yang and Raji{\v{c}}, Frano and Tang, Siyu and Wetzstein, Gordon},
  journal={arXiv preprint arXiv:2512.05076},
  year={2025}
}

@inproceedings{Li2025WorldModelBench,
  title={WorldModelBench: Judging Video Generation Models As World Models},
  author={Dacheng Li and Yunhao Fang and Yukang Chen and Shuo Yang and Shiyi Cao and Justin Wong and Michael Luo and Xiaolong Wang and Hongxu Yin and Joseph E. Gonzalez and Ion Stoica and Song Han and Yao Lu},
  booktitle = {Advances in Neural Information Processing Systems},
  year={2025},
}

@inproceedings{wu2025longtermmemory,
  title={Video world models with long-term spatial memory},
  author={Wu, Tong and Yang, Shuai and Po, Ryan and Xu, Yinghao and Liu, Ziwei and Lin, Dahua and Wetzstein, Gordon},
  booktitle = {Advances in Neural Information Processing Systems},
  year={2025},
}

@inproceedings{xiao2025worldmem,
  title={WorldMem: Long-term Consistent World Simulation with Memory},
  author={Xiao, Zeqi and Yushi, LAN and Zhou, Yifan and Ouyang, Wenqi and Yang, Shuai and Zeng, Yanhong and Pan, Xingang},
  booktitle={The Thirty-ninth Annual Conference on Neural Information Processing Systems},
  year={2025}
}

@article{zhou2025pai,
  title={PAI-Bench: A Comprehensive Benchmark For Physical AI},
  author={Zhou, Fengzhe and Huang, Jiannan and Li, Jialuo and Ramanan, Deva and Shi, Humphrey},
  journal={arXiv preprint arXiv:2512.01989},
  year={2025}
}

@article{rakheja2025worldconsistency,
  title={World Consistency Score: A Unified Metric for Video Generation Quality},
  author={Rakheja, Akshat and Ashdhir, Aarsh and Bhattacharjee, Aryan and Sharma, Vanshika},
  journal={arXiv preprint arXiv:2508.00144},
  year={2025}
}

@article{kwon2026toward,
  title={Toward Stable World Models: Measuring and Addressing World Instability in Generative Environments},
  author={Kwon, Soonwoo and Kim, Jin-Young and Go, Hyojun and Baek, Kyungjune},
  journal={Pattern Recognition},
  pages={113351},
  year={2026},
  publisher={Elsevier}
}

@inproceedings{li2025vmem,
  title={Vmem: Consistent interactive video scene generation with surfel-indexed view memory},
  author={Li, Runjia and Torr, Philip and Vedaldi, Andrea and Jakab, Tomas},
  booktitle={Proceedings of the IEEE/CVF International Conference on Computer Vision},
  pages={25690--25699},
  year={2025}
}

@inproceedings{li2025wonderplay,
  title={Wonderplay: Dynamic 3d scene generation from a single image and actions},
  author={Li, Zizhang and Yu, Hong-Xing and Liu, Wei and Yang, Yin and Herrmann, Charles and Wetzstein, Gordon and Wu, Jiajun},
  booktitle={Proceedings of the IEEE/CVF International Conference on Computer Vision},
  pages={9080--9090},
  year={2025}
}

@inproceedings{zheng2026versecrafter,
  title={VerseCrafter: Dynamic Realistic Video World Model with 4D Geometric Control},
  author={Zheng, Sixiao and Yin, Minghao and Hu, Wenbo and Li, Xiaoyu and Shan, Ying and Fu, Yanwei},
  booktitle={Proceedings of the IEEE/CVF Conference on Computer Vision and Pattern Recognition},
  year={2026}
}

@article{huang2026pointworld,
  title={PointWorld: Scaling 3D World Models for In-The-Wild Robotic Manipulation},
  author={Huang, Wenlong and Chao, Yu-Wei and Mousavian, Arsalan and Liu, Ming-Yu and Fox, Dieter and Mo, Kaichun and Fei-Fei, Li},
  journal={arXiv preprint arXiv:2601.03782},
  year={2026}
}

@inproceedings{zhou2025dinowm,
  title={DINO-WM: World Models on Pre-trained Visual Features enable Zero-shot Planning},
  author={Zhou, Gaoyue and Pan, Hengkai and LeCun, Yann and Pinto, Lerrel},
  booktitle={Forty-second International Conference on Machine Learning},
  year={2025}
}

@article{assran2025vjepa,
  title={V-jepa 2: Self-supervised video models enable understanding, prediction and planning},
  author={Assran, Mido and Bardes, Adrien and Fan, David and Garrido, Quentin and Howes, Russell and Muckley, Matthew and Rizvi, Ammar and Roberts, Claire and Sinha, Koustuv and Zholus, Artem and others},
  journal={arXiv preprint arXiv:2506.09985},
  year={2025}
}

@inproceedings{po2025long,
  title={Long-context state-space video world models},
  author={Po, Ryan and Nitzan, Yotam and Zhang, Richard and Chen, Berlin and Dao, Tri and Shechtman, Eli and Wetzstein, Gordon and Huang, Xun},
  booktitle={Proceedings of the IEEE/CVF International Conference on Computer Vision},
  pages={8733--8744},
  year={2025}
}

@article{lillemark2026flow,
  title={Flow Equivariant World Models: Memory for Partially Observed Dynamic Environments},
  author={Lillemark, Hansen Jin and Huang, Benhao and Zhan, Fangneng and Du, Yilun and Keller, Thomas Anderson},
  journal={arXiv preprint arXiv:2601.01075},
  year={2026}
}

@inproceedings{yang2025cogvideox,
  title={Cogvideox: Text-to-video diffusion models with an expert transformer},
  author={Yang, Zhuoyi and Teng, Jiayan and Zheng, Wendi and Ding, Ming and Huang, Shiyu and Xu, Jiazheng and Yang, Yuanming and Hong, Wenyi and Zhang, Xiaohan and Feng, Guanyu and others},
  booktitle={International Conference on Learning Representations},
  year={2025}
}

@inproceedings{ren2025gen3c,
  title={Gen3c: 3d-informed world-consistent video generation with precise camera control},
  author={Ren, Xuanchi and Shen, Tianchang and Huang, Jiahui and Ling, Huan and Lu, Yifan and Nimier-David, Merlin and M{\"u}ller, Thomas and Keller, Alexander and Fidler, Sanja and Gao, Jun},
  booktitle={Proceedings of the IEEE/CVF Conference on Computer Vision and Pattern Recognition},
  pages={6121--6132},
  year={2025}
}

@inproceedings{zhu2025aether,
  title={Aether: Geometric-aware unified world modeling},
  author={Zhu, Haoyi and Wang, Yifan and Zhou, Jianjun and Chang, Wenzheng and Zhou, Yang and Li, Zizun and Chen, Junyi and Shen, Chunhua and Pang, Jiangmiao and He, Tong},
  booktitle={Proceedings of the IEEE/CVF International Conference on Computer Vision},
  pages={8535--8546},
  year={2025}
}

@inproceedings{viswanathan2025checklist,
title = {Checklists Are Better Than Reward Models For Aligning Language Models},
booktitle = {NeurIPS},
author = {Vijay Viswanathan and Yanchao Sun and Shuang Ma and Xiang Kong and Meng Cao and Graham Neubig and Tongshuang Wu},
year = {2025},
URL = {https://arxiv.org/abs/2507.18624}
}

@article{li2025smallworlds,
  title={SmallWorlds: Assessing Dynamics Understanding of World Models in Isolated Environments},
  author={Li, Xinyi and Xia, Zaishuo and Lu, Weyl and Hao, Chenjie and Chen, Yubei},
  journal={arXiv preprint arXiv:2511.23465},
  year={2025}
}

@inproceedings{ling2024dl3dv,
  title={Dl3dv-10k: A large-scale scene dataset for deep learning-based 3d vision},
  author={Ling, Lu and Sheng, Yichen and Tu, Zhi and Zhao, Wentian and Xin, Cheng and Wan, Kun and Yu, Lantao and Guo, Qianyu and Yu, Zixun and Lu, Yawen and others},
  booktitle={Proceedings of the IEEE/CVF Conference on Computer Vision and Pattern Recognition},
  pages={22160--22169},
  year={2024}
}

@misc{li2025sekaivideodatasetworld,
      title={Sekai: A Video Dataset towards World Exploration}, 
      author={Zhen Li and Chuanhao Li and Xiaofeng Mao and Shaoheng Lin and Ming Li and Shitian Zhao and Zhaopan Xu and Xinyue Li and Yukang Feng and Jianwen Sun and Zizhen Li and Fanrui Zhang and Jiaxin Ai and Zhixiang Wang and Yuwei Wu and Tong He and Jiangmiao Pang and Yu Qiao and Yunde Jia and Kaipeng Zhang},
      year={2025},
      booktitle={Advances in Neural Information Processing Systems}
}

@article{zhao2025spatia,
  title={Spatia: Video Generation with Updatable Spatial Memory},
  author={Zhao, Jinjing and Wei, Fangyun and Liu, Zhening and Zhang, Hongyang and Xu, Chang and Lu, Yan},
  journal={arXiv preprint arXiv:2512.15716},
  year={2025}
}

@article{realestate10k,
title	= {Stereo Magnification: Learning view synthesis using multiplane images},
author	= {Tinghui Zhou and Richard Tucker and John Flynn and Graham Fyffe and Noah Snavely},
year	= {2018},
URL	= {https://arxiv.org/abs/1805.09817},
journal	= {ACM Trans. Graph. (Proc. SIGGRAPH)},
volume	= {37}}

@article{bansal2025videophy2,
  title={Videophy-2: A challenging action-centric physical commonsense evaluation in video generation},
  author={Bansal, Hritik and Peng, Clark and Bitton, Yonatan and Goldenberg, Roman and Grover, Aditya and Chang, Kai-Wei},
  journal={arXiv preprint arXiv:2503.06800},
  year={2025}
}

@inproceedings{bansal2024videophy,
  title={Videophy: Evaluating physical commonsense for video generation},
  author={Bansal, Hritik and Lin, Zongyu and Xie, Tianyi and Zong, Zeshun and Yarom, Michal and Bitton, Yonatan and Jiang, Chenfanfu and Sun, Yizhou and Chang, Kai-Wei and Grover, Aditya},
  booktitle={International Conference on Learning Representations},
  year={2025}
}
